\documentclass[journal]{IEEEtran}
\usepackage[usenames]{color}
\usepackage{amsmath,amssymb,graphicx,url, epsfig}
\usepackage{cite, subfigure, xspace, paralist}

\def \tstc {\texttt{tst2013}\xspace}
\def \tsta {\texttt{tst2010}\xspace}
\def \tstb {\texttt{tst2011}\xspace}
\def \dev {\texttt{dev2010}\xspace}
\def \amidev {\texttt{dev}\xspace}
\def \amieval {\texttt{eval}\xspace}
\def \swbdeval {\texttt{eval2000}\xspace}
\def \lhuc {\texttt{LHUC}\xspace}
\def \LHUC {\texttt{LHUC}\xspace}
\def \satlhuc {\texttt{SAT-LHUC}\xspace}

\def \R {\mathbb R}
\def \b #1{\textbf{#1}}

\def \f {\phi}
\def \nlf #1{\f^{#1}}

\def \C  {\mathcal{L}}

\def \b {b^{[L]}}

\def \x{{\mathbf x}}
\def \siparams{\mathbf{\theta}_{SI}}
\def \sdparams{\mathbf{\theta}_{SD}}
\def \sdlhuc{\mathbf{\theta}^{m_t}_{LHUC}}

\renewcommand{\vec}[1]{\mathbf{#1}}

\begin{document}

\title{Learning Hidden Unit Contributions for Unsupervised Acoustic Model Adaptation}

\author{Pawel~Swietojanski,~\IEEEmembership{Student Member,~IEEE,}\thanks{P Swietojanski and S Renals are with the Centre for Speech Technology Research, University of Edinburgh, Edinburgh EH89AB, U.K., email: \{p.swietojanski,s.renals\}@ed.ac.uk}
  Jinyu~Li,~\IEEEmembership{Member,~IEEE,}\thanks{J Li is with Microsoft Corporation. One Microsoft Way, WA, USA, email: jinyu.li@microsoft.com}
  and~Steve~Renals,~\IEEEmembership{Fellow,~IEEE}\thanks{PS and SR were supported by EPSRC Programme Grant grant EP/I031022/1 \emph{Natural Speech Technology} (NST) and the European Union under H2020 project \emph{SUMMA}, grant agreement 688139. The NST research data collection may be accessed at http://datashare.is.ed.ac.uk/handle/10283/786. This research utilised the K40 GPGPU board donated by NVIDA Corporation.}
}



%

\IEEEpubid{10.1109/TASLP.2016.2560534~\copyright~2016 IEEE.
}
\markboth{IEEE/ACM Transactions on Audio, Speech and Language Processing, Vol. 24, Num.8}{Swietojanski}

\maketitle

\begin{abstract}
This work presents a broad study on the adaptation of neural network acoustic models by means of learning hidden unit contributions (LHUC) -- a method that linearly re-combines hidden units in a speaker- or environment-dependent manner using small amounts of unsupervised adaptation data. We also extend LHUC to a speaker adaptive training (SAT) framework that leads to a more adaptable DNN acoustic model, working both in a speaker-dependent and a speaker-independent manner, without the requirements to maintain auxiliary speaker-dependent feature extractors or to introduce significant speaker-dependent changes to the DNN structure. Through a series of experiments on four different speech recognition benchmarks (TED talks, Switchboard, AMI meetings, and Aurora4) comprising 270 test speakers, we show that LHUC in both its test-only and SAT variants results in consistent word error rate reductions ranging from 5\% to 23\% relative depending on the task and the degree of mismatch between training and test data.  In addition, we have investigated the effect of the amount of adaptation data per speaker, the quality of unsupervised adaptation targets, the complementarity to other adaptation techniques, one-shot adaptation, and an extension to adapting DNNs trained in a sequence discriminative manner.
\end{abstract}



%

\section{Introduction and Summary}

\IEEEPARstart{S}{peech} recognition accuracies have improved substantially  over the past several years through the use of (deep) neural network (DNN) acoustic models. Hinton et al \cite{Hinton2012} report word error rate (WER) reductions between 10--32\% across a wide variety of tasks, compared with discriminatively trained 
Gaussian mixture model (GMM) based systems.  These results use neural networks as part of both hybrid DNN/HMM (hidden Markov model) systems~\cite{Bourlard1994,Renals1994, Seide2011, Dahl2012, Hinton2012} in which the neural network provides a scaled likelihood estimate to replace the GMM, and as tandem or bottleneck feature systems~\cite{Hermansky2000,Grezl2007} in which the neural network is used as a discriminative feature extractor for a GMM-based system.  For many tasks it has been observed that GMM-based systems (with tandem or bottleneck features) that have been adapted to the talker are more accurate than unadapted hybrid DNN/HMM systems~\cite{Sainath2013_cnns, Swietojanski2013, Woodland2015}, indicating that the adaptation of DNN acoustic models is an important topic that merits investigation.

Acoustic model adaptation~\cite{woodland2001} aims to normalise the mismatch between training and runtime data distributions that arises owing to the acoustic variability across speakers, as well as other distortions introduced by the channel or acoustic environment. 
In this paper we investigate unsupervised model-based adaptation of DNN acoustic models to speakers and to acoustic environments, using a recently introduced method called  \emph{Learning Hidden Unit Contributions} (\LHUC)~\cite{Abdel-Hamid2013_is, Swietojanski2014_lhuc, Swietojanski:ICASSP16}.  We present the \LHUC approach both in the context of test-only adaptation, and an extension to speaker-adaptive training (SAT), referred to as \satlhuc~\cite{Swietojanski:ICASSP16}.  We present an extensive experimental analysis using four standard corpora: TED talks~\cite{Cettolo2012}, AMI~\cite{Carletta_LRE2007}, Switchboard~\cite{godfrey1992switchboard} and Aurora4~\cite{aurora4}.  These experiments include:  adaptation of both cross-entropy and sequence trained DNN acoustic models (Sec.~\ref{ssec:base}--\ref{ssec:sequence}); an analysis in terms of the quality of adaptation targets, quality of adaptation data and the amount of adaptation data (Sec.~\ref{ssec:quality}); complementarity with feature-space adaptation techniques based on maximum likelihood linear regression~\cite{Gales1998} (Sec.~\ref{ssec:complementarity}); and application to combined speaker and environment adaptation (Sec.~\ref{sec:factorisation}).
\IEEEpubidadjcol 
\section{Review of Neural Network Acoustic Adaptation} \label{ssec:adapt}
Approaches to the adaptation of neural network acoustic models can be considered as operating either in the feature space, or in the model space, or as a hybrid approach in which speaker-, utterance-, or environment-dependent auxiliary features are appended to the standard acoustic features.

The dominant technique for estimating \emph{feature space transforms} is constrained (feature-space) MLLR, referred to as CMLLR or fMLLR~\cite{Gales1998}.  fMLLR is an adaptation method developed for GMM-based acoustic models, in which an affine transform of the input acoustic features is estimated by maximising the log-likelihood that the model generates the adaptation data based on first pass alignments. To use fMLLR with a DNN-based system, it is first necessary to train a complete GMM-based system, which is then used to estimate a single input transform per speaker.  The transformed feature vectors are then used to train a DNN in a speaker adaptive manner and another set of transforms is estimated (using the GMM) during evaluation for unseen speakers. This technique has been shown to be effective in reducing WER across several different data sets, in both hybrid and tandem approaches~\cite{Mohamed2011, Seide2011, Hain2012, Hinton2012, Sainath2012, Sainath2013_cnns, Bell2013_mlan, Swietojanski2013}.  Similar techniques have also been developed to operate directly on neural networks. The linear input network (LIN)~\cite{Neto1995, Abrash1995} defines an additional speaker-dependent layer between the input features and the first hidden layer, and thus has a similar effect to fMLLR.  This technique has been further developed to include the use of a tied variant of LIN in which each of the input frames is constrained to have the same linear transform~\cite{li2010,Seide2011}. LIN and tied-LIN have been mostly used in test-only adaptation schemes; to make use of fMLLR transforms one needs to perform SAT training, which can usually better compensate against variability in acoustic space. 

An alternative speaker-adaptive training approach -- \emph{auxiliary features} -- augments the acoustic feature vectors with additional speaker-specific features computed for each speaker at both training and test stages.  There has been considerable recent work exploring the use of i-vectors~\cite{Dehak2010} for this purpose.  I-vectors, which can be regarded as basis vectors which span a subspace of speaker variability, were first used for adaptation in a GMM framework by Karafiat et al~\cite{Karafiat2011}.  Saon et al~\cite{Saon2013} used i-vectors to augment the input features of DNN-based  acoustic models, and showed that appending 100-dimensional i-vectors for each speaker resulted in a 10\% relative reduction in WER on Switchboard (and a 6\% reduction when the input features had been transformed using fMLLR). Gupta et al~\cite{Gupta2014} obtained similar results, and Karanasou et al~\cite{Karanasou2014} presented an approach in which the i-vectors were factorised into speaker and environment parts. Miao et al~\cite{Miao2015} proposed to transform i-vectors using an auxiliary DNN which produced speaker-specific transforms of the original feature vectors, similar to fMLLR. Other examples of auxiliary features include the use of speaker-specific bottleneck features obtained from a speaker separation DNN used in a distant speech recognition task~\cite{Liu2014}, the use of out-of-domain tandem features~\cite{Bell2013_mlan}, and speaker codes~\cite{Bridle1990, Abdel-Hamid2013, Xue2014_scodes} in which a specific set of units for each speaker is optimised. Speaker codes  require speaker adaptive (re-)training, owing to the additional connection weights between codes and hidden units.

\emph{Model-based adaptation} relies on a direct update of DNN parameters.  Liao~\cite{Liao2013}  investigated supervised and unsupervised adaptation of different weight subsets using a few minutes of adaptation data.  On a large net (60M weights), up to 5\% relative improvement was observed for unsupervised adaptation when all weights were adapted.  Yu et al~\cite{Yu2013} have explored the use of regularisation for adapting the weights of a DNN,  using the Kullback-Liebler (KL) divergence between the speaker-independent (SI) and speaker-dependent (SD) output distributions, resulting in a 3\% relative improvement on Switchboard.  This approach was also recently used to adapt all parameters of sequence-trained models~\cite{Huang2015}.  One can also reduce the number of speaker-specific parameters through a different forms of factorisation~\cite{Xue2014, samarakoon2015learning}.  Ochiai et al~\cite{Ochiai2014} have also explored regularised speaker adaptive training with a speaker-dependent layer.

Directly adapting the weights of a large DNN results in extremely large speaker-dependent parameter sets, and a computationally intensive adaptation process.  Smaller subsets of the DNN weights may be modified, including output layer biases \cite{Yao2012}, the bias and slope of hidden units~\cite{Zhao2015_slopes} or training the models with differentiable pooling operators~\cite{Swietojanski:ICASSP15}, which are then adapted in SD fashion. Siniscalchi et al \cite{Sini2013} also investigated the use of Hermite polynomial activation functions, whose parameters are estimated in a speaker adaptive fashion.  One can also adapt the top layer in a Bayesian fashion resulting in a maximum a posteriori (MAP) approach~\cite{huang2015maximum}, or address the sparsity of context-dependent tied-states when few adaptation data-points are available by using multi-task adaptation, using monophones to adapt the context-dependent output layer~\cite{Huang2015_mt, Swietojanski2015_mt}. A similar approach, but using a hierarchical output layer (tied-states followed by monophones) rather than multi-task adaptation, has also been proposed~\cite{Price2014}.

\section{Learning Hidden Unit Contributions (LHUC)} \label{sec:lhuc}

A neural network may be viewed as a set of \emph{adaptive basis functions}. Under certain assumptions on the family of target functions $f^*$ (as well as on the model structure itself) the neural network can act as an universal approximator~\cite{hornik1989multilayer, Hornik1991, Barron1993}. That is, given some vector of input random variables $\vec x \in \R^{d}$ there exists a neural network $f_n(\vec x) : \R^{d} \rightarrow \R$ of the form
\begin{equation} \label{eq:basis}
f_n(\vec x) = \sum_{k=1}^n r_k\psi(\vec w^\top_k\vec x + b_k) 
\end{equation}
which can approximate $f^*$ with an arbitrarily small error $\epsilon$ with respect to a distance measure such as mean square error (provided $n$ is sufficiently large):
\begin{equation}
||f^*(\vec x)-f_n(\vec x)||_2 \leq \epsilon .
\end{equation}
In~\eqref{eq:basis} $\psi : \R \rightarrow \R$ is an element-wise non-linear operation applied after an affine transformation which forms an adaptive basis function parametrised by a set of biases $b_k \in \R$ and a weight vectors $\vec w_k\in\R^{d_{\vec x}}$. The target approximation may then be constructed as a linear combination of the basis functions, each weighted by $r_k\in \R$. The formulation can be extended to $m$-dimensional mappings $f^m_n(\vec x) : \R^{d} \rightarrow \R^m$ simply by splicing the models in~\eqref{eq:basis} $m$ times.  The properties also hold true when considering deeper (nested) models \cite{hornik1989multilayer} (Corollaries 2.6 and 2.7).

\begin{figure}[t]
\center
  \includegraphics[width=0.95\columnwidth]{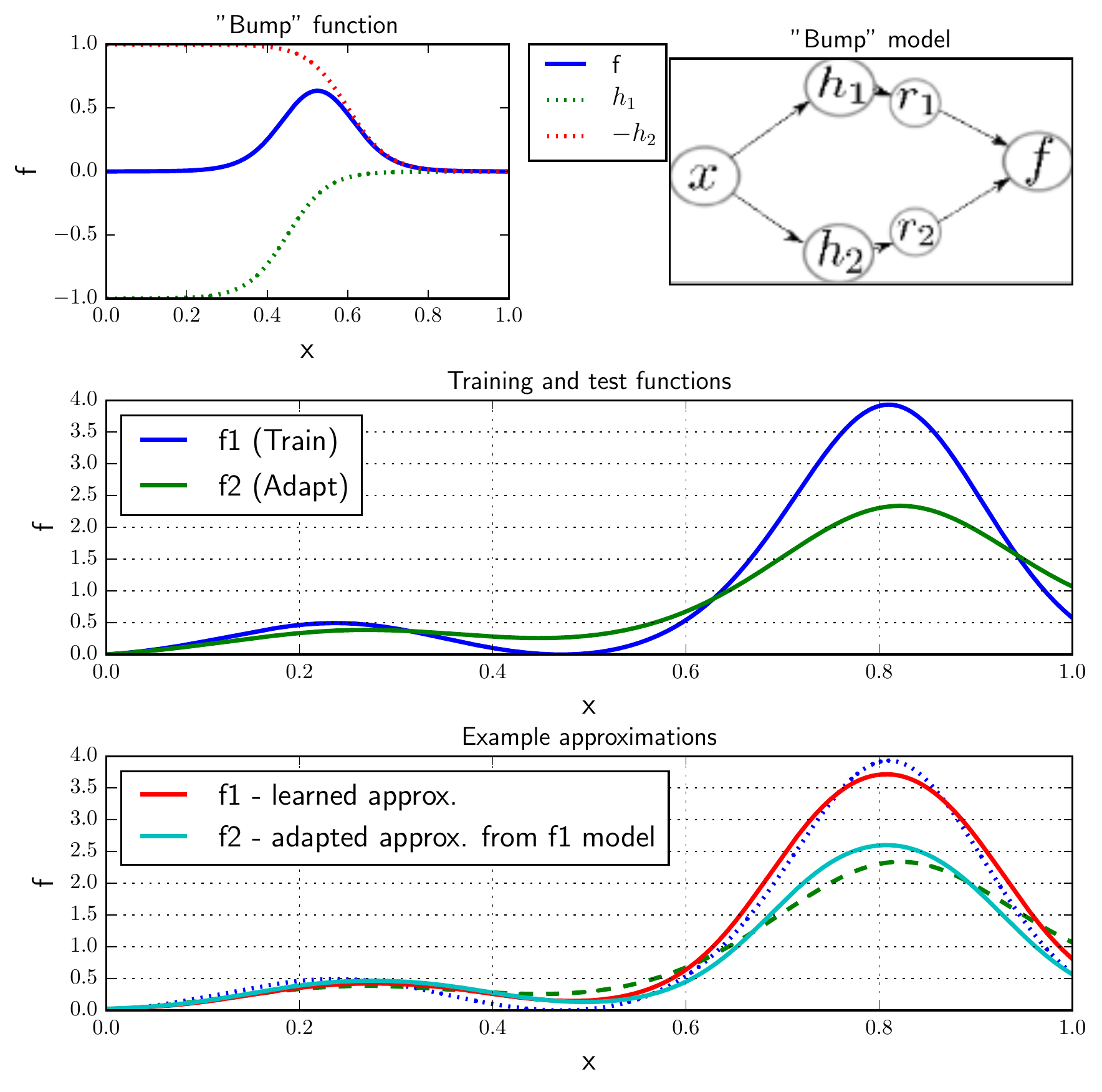}
\caption{Example illustration of how \LHUC performs adaptation (best viewed in color). Top: A ``bump'' model (eq. \ref{eq:basis}) with two hidden units can approximate ``bump'' functions. Middle: To learn function $f_2$ given training data $f_1$ (middle), we splice two ``bump'' functions together (4 hidden units, one input/output) to learn an approximation of function $f_1$. Bottom: \LHUC adaptation of the model optimised to $f_1$ and adapted to $f_2$ using \LHUC scaling parameters. Image reproduced from \cite{Swietojanski:ICASSP16}.}
\label{fig:lhuc_si}
\end{figure}

DNN training results in the hidden units learning a joint representation of the target function and becoming specialised and complementary to each other.  Generalisation corresponds to the learned combination of basis functions continuing to approximate the target function when applied to unseen test data.  This interpretation motivates the idea of using \LHUC -- Learning Hidden Unit Contributions -- for test-set adaptation.  In \LHUC the network's basis functions, previously estimated using a large amount of training data, are kept fixed.  Adaptation involves modifying the combination of hidden units in order to minimise the adaptation loss based on the adaptation data.  Fig.~\ref{fig:lhuc_si}  illustrates this approach for a regression problem, where the adaptation is performed by linear re-combination of basis functions changing only the $r$ parameters from eq.~\eqref{eq:basis}.

The key idea of \LHUC is to explicitly parametrise the amplitudes of each hidden unit (either in fully-connected and convolutional layers after max-pooling), using a speaker-dependent amplitude function.  Let $h^{l,s}_j$ denote the $j$-th hidden unit activation (basis) in layer $l$, and let $r^{l,s}_j \in \R{}$ denote the $s$-th speaker-dependent amplitude function:
\begin{align} \label{eq:lhuc}
  h^{l,s}_j = \xi(r^{l,s}_j)\circ \psi_j \left (\vec w^{l\top}_{j} \vec x + b^l_j \right) .
\end{align}
The amplitude is modelled using a function $\xi : \R \rightarrow \R^+$  -- typically a sigmoid with range $(0,2)$~\cite{Swietojanski2014_lhuc}, but an identity function could be used~\cite{Zhang2015}. $\vec w^l_j$ is the $j$th column of the corresponding weight matrix $\vec W^l$, $b_j^l$ denotes the bias, $\psi$ is the hidden unit activation function (unless stated otherwise, this is assumed to be sigmoid), and $\circ$ denotes a Hadamard product\footnote{Although the equations are given in scalar form, we have used Hadamard product notation to emphasise the operation that would be performed once expanded to full-rank matrices.}.  $\xi$ constrains the  range of the hidden unit amplitude scaling (compare with Fig.~\ref{fig:lhuc_si}) hence directly affecting the adaptation transform capacity -- this may be desirable when adapting with potentially noisy unsupervised targets (see Sec.~\ref{ssec:base}).  \LHUC adaptation progresses by setting the speaker-specific amplitude parameters $r^{l,s}_j$ using gradient descent with targets provided by the adaptation data.

 The idea of directly learning hidden unit amplitudes was proposed in the context of an adaptive learning rate schedule by Trentin~\cite{Trentin2001}, and was later applied to supervised speaker adaptation by Abdel-Hamid and Jiang~\cite{Abdel-Hamid2013_is}. The approach was extended to unsupervised adaptation, non-sigmoid non-linearities, and large vocabulary speech recognition by Swietojanski and Renals~\cite{Swietojanski2014_lhuc}. Other adaptive transfer function methods for speaker adaptation have also been proposed~\cite{Sini2013,Zhao2015_slopes}, as have ``basis'' techniques~\cite{Wu2015, Tian2015, Decloirx2015}. However, the basis in the latter works involved re-tuning parallel models on pre-defined clusters (gender, speaker, environment) in a supervised manner; the adaptation then relied on learning linear combination coefficients for those sub-models on adaptation data.

\begin{figure}[t]
\center
  \includegraphics[width=0.95\columnwidth]{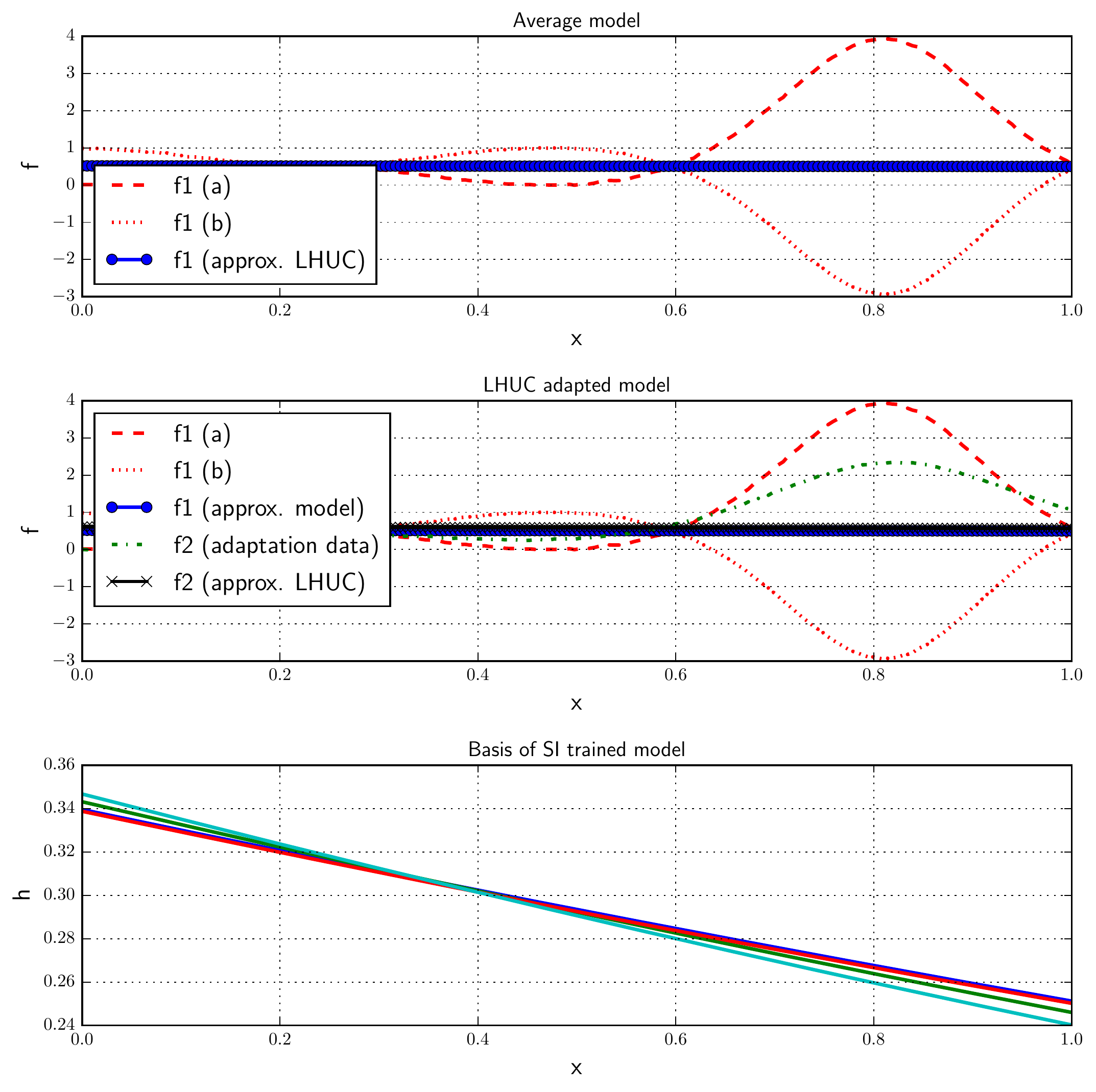}
\caption{A 4-hidden-unit model trained to $f1 (a)$ and $f1 (b))$ for an SI approach (top) and an adapted representation (middle) keeping the resulting basis functions fixed (bottom).  (Best viewed in color.)}
\label{fig:lhuc_basis}
\end{figure}

\section{Speaker Adaptive Training LHUC (SAT-LHUC)}  \label{sec:satlhuc}
When \LHUC is applied as a test-only adaptation it assumes that the set of speaker-independent basis functions estimated on the training data provides a good starting point for further tuning to the underlying data distribution of the adaptation data (Fig.~\ref{fig:lhuc_si}). However, one can derive a counter-example where this assumption fails: the top plot of Fig.~\ref{fig:lhuc_basis} shows example training data uniformly drawn from two competing distributions $f1(a)$ and $f1(b)$ where the linear recombination of the resulting basis in the average model (Fig~\ref{fig:lhuc_basis} bottom), provides a poor approximation of adaptation data.
 
\begin{figure}[t]
\center
  \includegraphics[width=0.95\columnwidth]{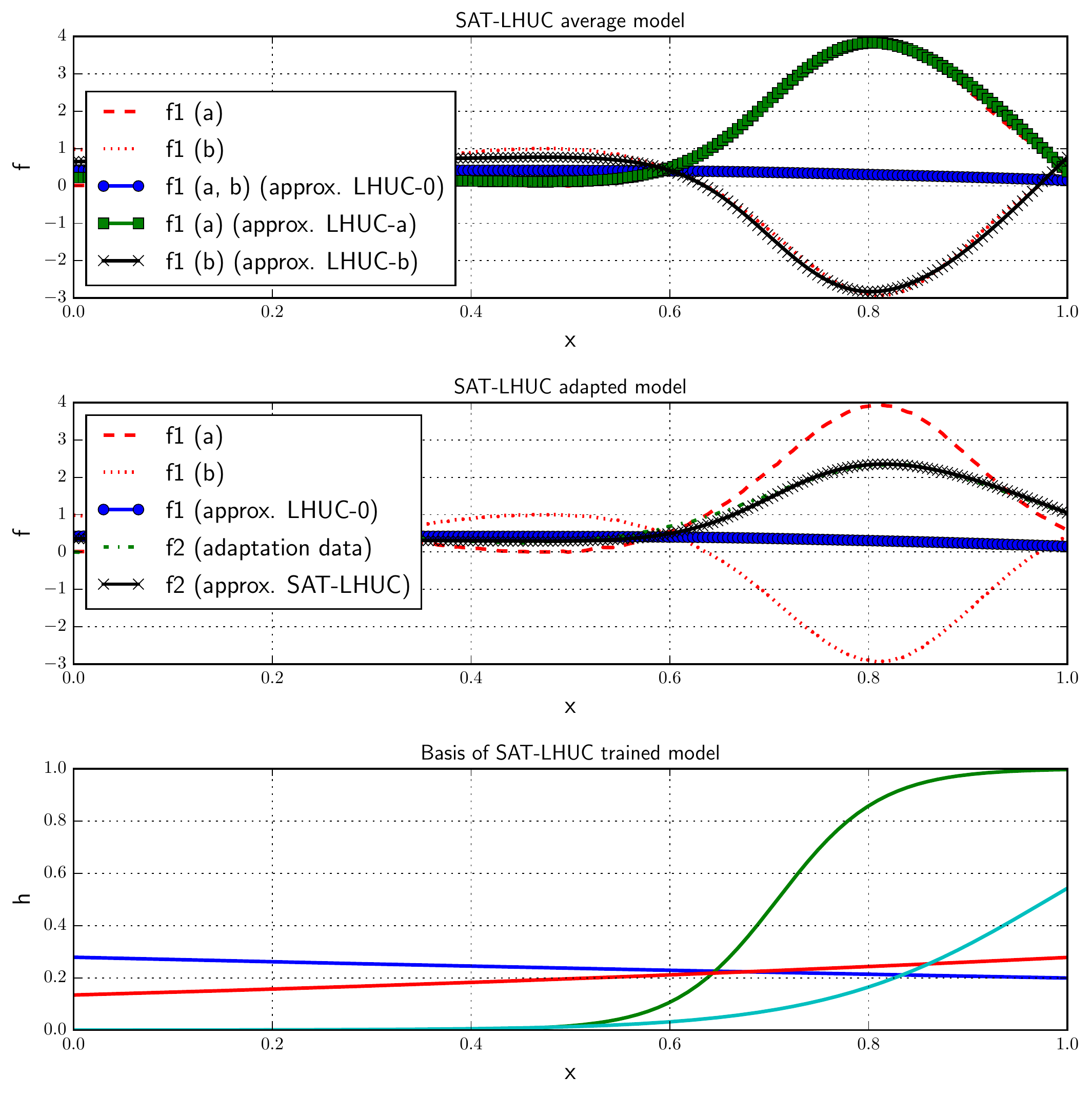}
\caption{Learned solutions using three different \satlhuc transforms  and shared basis functions: \LHUC-0 learns to provide a good average fit to both distributions $f1 (a)$ and $f1 (b))$ at the same time, while \LHUC-a and \LHUC-b are tasked to fit either $f1 (a)$ or $f1 (b)$, respectively. The bottom  plot shows the resulting basis functions (activations of 4 hidden units) of the \satlhuc training approach - one can observe \satlhuc provides a richer set of basis function which can fit  the data well on average, and can also capture some underlying characteristics necessary to reconstruct target training data -- using different \LHUC transforms, this property is also visualised in the middle plot. (Best viewed in color.)}
\label{fig:satlhuc_basis}
\end{figure}

This motivates combining \LHUC with speaker adaptive training (SAT)~\cite{Anastasakos1996} in which the hidden units are trained to capture both good average representations and speaker-specific representations, by estimating speaker-specific hidden unit amplitudes for each training speaker.  This is visualised in Fig.~\ref{fig:satlhuc_basis} where, given the prior knowledge of which data-point comes from which distribution, we estimate a set of parallel \lhuc transforms (one per distribution) as well as one extra transform which is responsible for modelling average properties. The top of Fig.~\ref{fig:satlhuc_basis} shows the same experiment as in Fig~\ref{fig:lhuc_basis} but with three \lhuc transforms -- one can see that the 4-hidden-unit MLP in this scenario was able to capture each of the underlying distributions as well as the average aspect well, given the \lhuc transform. At the same time, the resulting basis functions (Fig~\ref{fig:satlhuc_basis}, bottom) are a better starting point for the adaptation (Fig.~\ref{fig:satlhuc_basis}, middle).

The examples presented in Figs.~\ref{fig:lhuc_basis} and~\ref{fig:satlhuc_basis} could be solved by breaking the symmetry through rebalancing the number of  training data-points for each function, resulting in less trivial and hence more adaptable basis functions in the average model. However, as we will show experimentally later,  similar effects are also present in high-dimensional speech data, and  \satlhuc training allows more tunable canonical acoustic models to be built, that can be better tailored to particular speakers through adaptation.

Test-only adaptation for \satlhuc remains the same as for \lhuc -- the set of speaker-dependent \lhuc parameters 
$ \theta^s_{LHUC} = \{r^{l,s}_j\}$ 
is inserted for each test speaker and their values optimised from unsupervised adaptation data. We also use a set of \lhuc transforms $\theta^s_{LHUC}$, where $s=1\ldots S$,  for the training speakers  which are jointly optimised with the speaker-independent parameters  
$\theta_{SI}=\{\vec W^l, \vec b^l\}$. 
There is an additional speaker-independent LHUC transform, denoted by $\theta^0_{LHUC}$, which allows the model to be used in speaker-independent fashion, for example, to produce first pass adaptation targets. This joint learning process of hidden units with speaker-dependent \lhuc scalers is important, as it results in a more tunable canonical acoustic model that can be better adjusted to unseen speakers at test time, as we have illustrated in Fig.~\ref{fig:satlhuc_basis} and demonstrated on adaptation tasks in the following sections.

\begin{figure}
\center
\includegraphics[width=1.0\columnwidth]{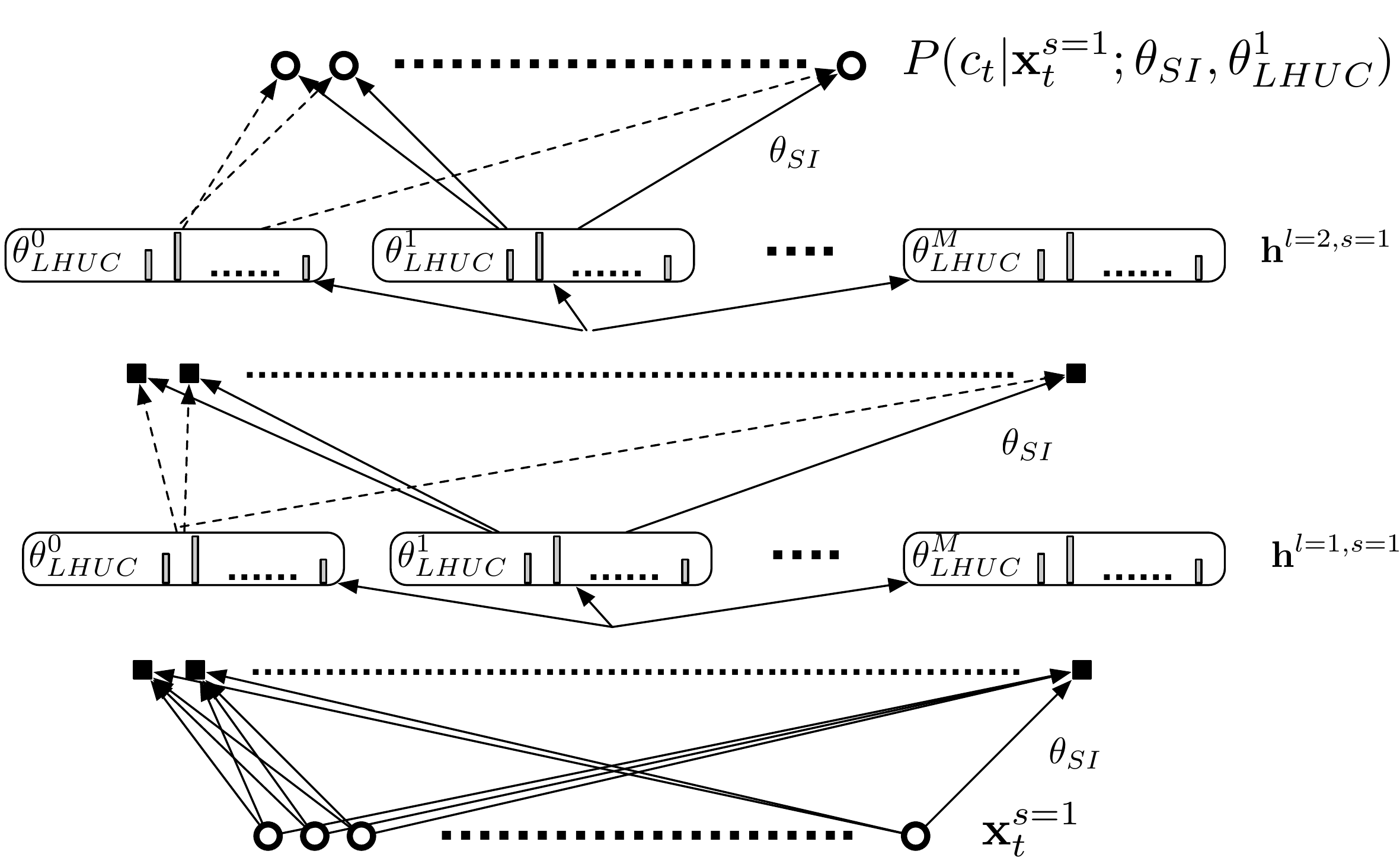}
\caption{Schematic of SAT-LHUC training, with a data point from speaker $s=1$. Dashed line indicates an alternative route through the SI LHUC transform.}
\label{fig:sat}
\end{figure}

To perform SAT training with \LHUC, we use the negative log likelihood and maximise the posterior probability of obtaining the correct context-dependent tied-state $c_t$ given observation vector $\vec x_t$ at time $t$:
\begin{align}
  \C_{SAT}&(\siparams,\sdparams) = -\sum_{t \in D} \log P(c_t|\x^{s}_t;\siparams;\sdlhuc)
  \label{eq:sat_lhuc}
\end{align}
where $s$ denotes the $s$th speaker, $m_t\in\{0,s\}$ selects the SI or SD \LHUC transforms from $\theta_{SD}\in\{\theta^0_{LHUC},\ldots,\theta^S_{LHUC}\}$ based on a Bernoulli distribution:
\begin{align}
k_t &\sim\ \mbox{Bernoulli}(\gamma) \label{eq:binomial1}\\
m_t &=
\begin{cases}
    s       & \quad \text{if } k_t = 0 \\
    0       & \quad \text{if } k_t = 1 \\
\end{cases}
\label{eq:binomial}
\end{align}
where $\gamma$ is a hyper-parameter specifying the probability the given example is treated as SI. The SI/SD split (determined by equations~\eqref{eq:binomial1} and~\eqref{eq:binomial}) can be performed at speaker, utterance or frame level. We further investigate this aspect in section ~\ref{ssec:sat}. The \satlhuc model structure is depicted in Fig~\ref{fig:sat}; notice the alternative routes of forward and backward passes for different speakers.

Denote by $\partial \C_{SAT} / \partial h^{l,s}_j$ the error back-propagated to the $j$th unit at the $l$th layer (eq.~\eqref{eq:lhuc}). To back propagate through the transform one needs to element-wise multiply it by the transform itself, as follows:
\begin{equation}
   \frac{\partial \C_{SAT}}{\partial \psi^l_j} = \frac{\partial \C_{SAT}}{\partial h^{l,s}_j} \circ \xi(r^{l,s}_j) \, .
\end{equation}
To obtain the gradient with respect to $r^{l,s}_j$:
\begin{equation}
   \frac{\partial \C_{SAT}}{\partial r^{l,s}_j} = \frac{\partial \C_{SAT}}{\partial h^{l,s}_j} \circ \frac{\partial \xi(r^{l,s}_j)}{\partial r^{l,s}_j} \circ \psi^l_j \, .
\end{equation}
When performing mini-batch SAT training one needs to explicitly take account of the fact that different data-points may flow through different transforms: hence the resulting gradient for $r^{l,s}_j$ for the $s$th speaker is the sum of the partial gradients belonging to speaker $s$:
\begin{equation}
   \frac{\partial \C_{SAT}}{\partial r^{l,s}_j} = \sum_{t,m_t=s}\frac{\partial \C_{SAT}}{\partial h^{l,s}_j}\circ\frac{\partial \xi(r^{l,s}_j)}{\partial r^{l,s}_j} \circ \psi^l_j \, ,
\end{equation}
or 0 in case no data-points for $s$th speaker in the given mini-batch were selected. All adaptation methods studied in this paper require first-pass decoding to obtain adaptation targets to either estimate fMLLR transforms for unseen test speakers or to perform DNN speaker-dependent parameter update. 

\section{Experimental setups}  \label{sec:setups}

We experimentally investigated \LHUC and \satlhuc using four different corpora: the TED talks corpus~\cite{Cettolo2012} following the IWSLT evaluation protocol (\url{www.iwslt.org}); the Switchboard corpus of conversational telephone speech~\cite{godfrey1992switchboard} (\url{ldc.upenn.edu}); the AMI meetings corpus \cite{Carletta_LRE2007, Renals_ASRU2007} (\url{corpus.amiproject.org}); and the Aurora4 corpus of read speech with artificially corrupted acoustic environments~\cite{aurora4} (\url{catalog.elra.info}). Unless explicitly stated otherwise, the models share similar structure across the tasks -- DNNs with 6 hidden layers (2,048 units in each) using a sigmoid non-linearity.  The output logistic regression layer models the distribution of context-dependent clustered tied states~\cite{Dahl2012}. The features are presented in 11 ($\pm5$) frame long context windows. All the adaptation experiments, unless explicitly stated otherwise, were performed unsupervised.

Below, we briefly describe each of the above corpora and its specific experimental configurations. The collective summary of adaptation-related statistics for each corpora is given in Table~\ref{tab:corpora}.  Note that we adapt to the headset or the side of a conversation, rather than the actual speaker (unless stated otherwise). As a result, the actual number of clusters (or estimated transforms) during training may differ from the number of physical speakers in the data.

\begin{table}[t]
\small
\caption{Corpus statistics related to SAT and adaptation. In parentheses we give the actual number of speakers.}
\label{tab:corpora}
\centerline{
\begin{tabular}{l|c|c||c|c}
& \multicolumn{2}{c||}{Training} & \multicolumn{2}{c}{Test}\\ \cline{2-5}
Corpora & \#Clusters & Time (h) & \#Clusters & Time (h)\\ \hline \hline
Aurora4 & 83 (83) & 15 & 8 (8) &  8.8 \\
AMI & 547 (155) & 80 & 135 (36) & 17.5\\
TED & 788 (788) & 143 & 39 (39) & 9.0 \\
SWBD & 4804 (4000) & 283 & 80 (80) & 3.6\\
\hline\hline
\end{tabular}}
\end{table}

\textbf{TED}: We carried out experiments using a corpus of publicity available TED talks (\url{www.ted.com})  following the IWSLT ASR evaluation protocol~\cite{Federico2012_iwslt} (\url{iwslt.org}). The training data consisted of 143 hours of speech (813 talks) and the systems follow our previously described recipe~\cite{Swietojanski2013}. In this work however, compared to our previous works~\cite{Swietojanski2013, Swietojanski2014_lhuc, Swietojanski:ICASSP15}, our systems employ more accurate language models developed for our IWSLT--2014 systems~\cite{Bell2014}: in particular, the final reported results use a 4-gram language model estimated from 751~million words. The baseline TED acoustic models are trained on unadapted PLP features with first and second order time derivatives. We present results on four predefined IWSLT test sets:  \dev, \tsta, \tstb and \tstc containing 8, 11, 8 and 28 ten-minute talks respectively. We use \tsta and/or \tstc to perform more detailed analyses. A collective summary of results on all TED test-sets is reported in Sec.~\ref{ssec:summary}. 

\textbf{AMI}: We follow the Kaldi GMM recipe described in~\cite{Swietojanski_ASRU13} and use acoustics from either Individual Headset Microphone (IHM) or Single Distant Microphone (SDM). In addition to cepstral features, we also trained a separate set of models using 40 mel-filter-bank (FBANK) features for which fMLLR transforms cannot be easily obtained (though not impossible~\cite{sainath2013fmllrcnn}), and for which \LHUC offers an interesting adaptation alternative. We also evaluated the effectiveness of \lhuc and \satlhuc applied to convolutional networks~\cite{Lecun1998, Abdel-Hamid2012, Sainath2013_cnns}, trained as described in~\cite{Swietojanski:SPL14} but with 300 convolutional filters. We decoded with a pruned 3-gram language model estimated from 800k words of AMI training transcripts interpolated with an LM trained on Fisher conversational telephone speech transcripts (1M words).

\textbf{Switchboard}: We use the Kaldi GMM recipe~\cite{Vesely:IS13, Kaldi:ASRU11}, using Switchboard--1 Release 2 (LDC97S62). Our baseline unadapted acoustic models were trained on LDA/MLLT features. The results are reported on the full Hub5 ’00 set (LDC2002S09) to which we will refer as \swbdeval. The \swbdeval contains two types of data,  Switchboard (SWBD) -- which is better matched to the training data -- and CallHome English (CHE). Our reported results use 3-gram LMs estimated from Switchboard and Fisher data.

\textbf{Aurora4}: The Aurora 4 task is a small scale, medium vocabulary noise and channel ASR robustness task based on the Wall Street Journal corpus \cite{aurora4}. We train our ASR models using the multi-condition training set. One half of the training utterances were recorded using a primary Sennheiser microphone, and the other half was collected using one of 18 other secondary microphones. The multi-condition set contains noisy utterances corrupted with one of six different noise types (airport, babble, car, restaurant, street traffic and train station) at 10-20 dB SNR. The standard Aurora 4 test set (\texttt{eval92}) consists of 330 utterances, which are used in 14 test conditions (4620 utterances in total). The same six noise types used during training are used to create noisy test utterances with SNRs ranging from 5-15dB SNR, resulting in a total of 14 test sets. These test sets are commonly grouped into 4 subsets -- clean (group A, 1 test case), noisy (group B, 6 test cases), clean with channel distortion (group C, 1 test case) and noisy with channel distortion (group D, 6 test cases). We decode with the standard task's 5k words bigram LM.

\section{Results} \label{sec:results}

\subsection{LHUC hyperparameters} \label{ssec:base}

\begin{figure*}[t]
\subfigure[]{
  \includegraphics[width=0.33\textwidth]{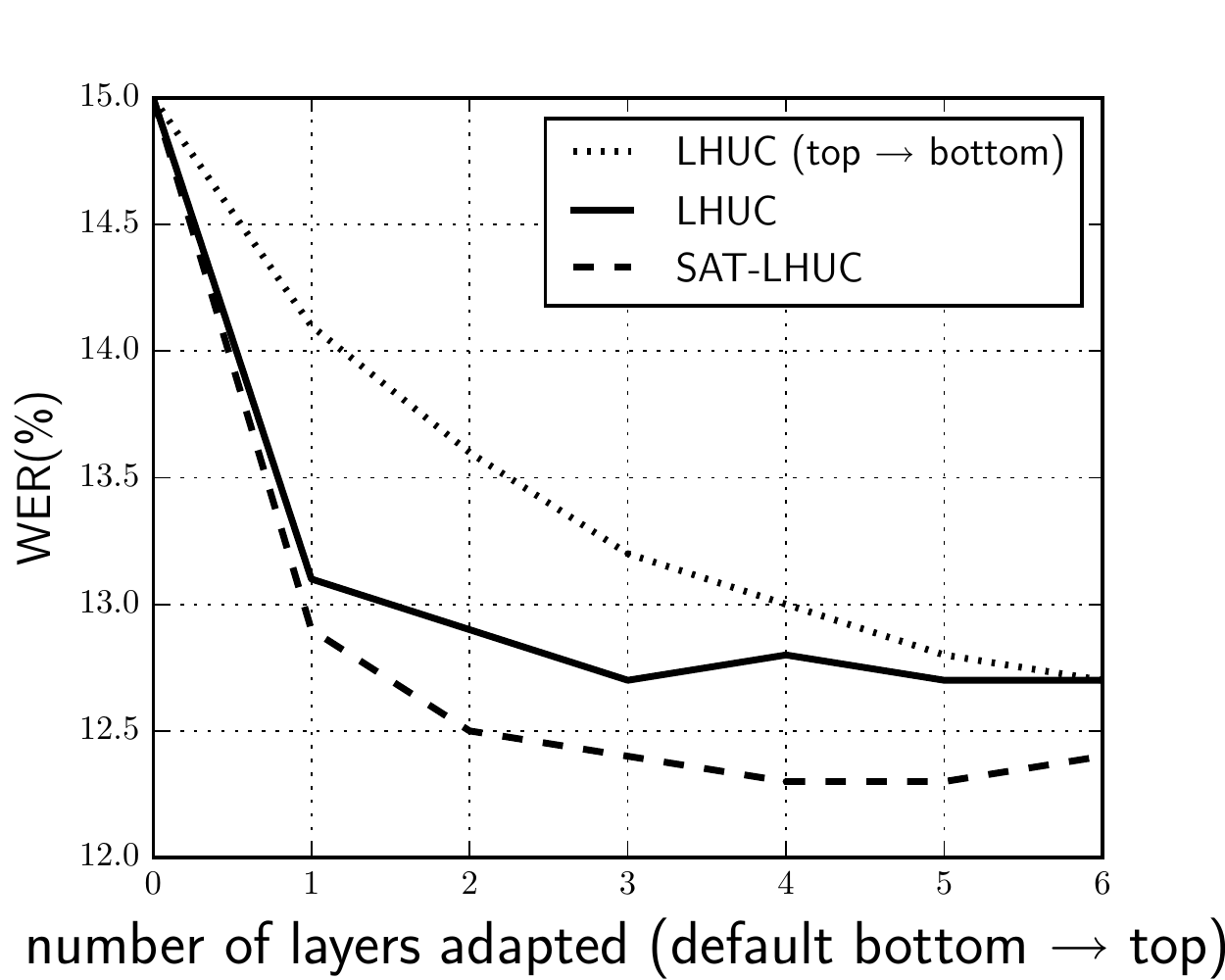}
}
\subfigure[]{
  \includegraphics[width=0.33\textwidth]{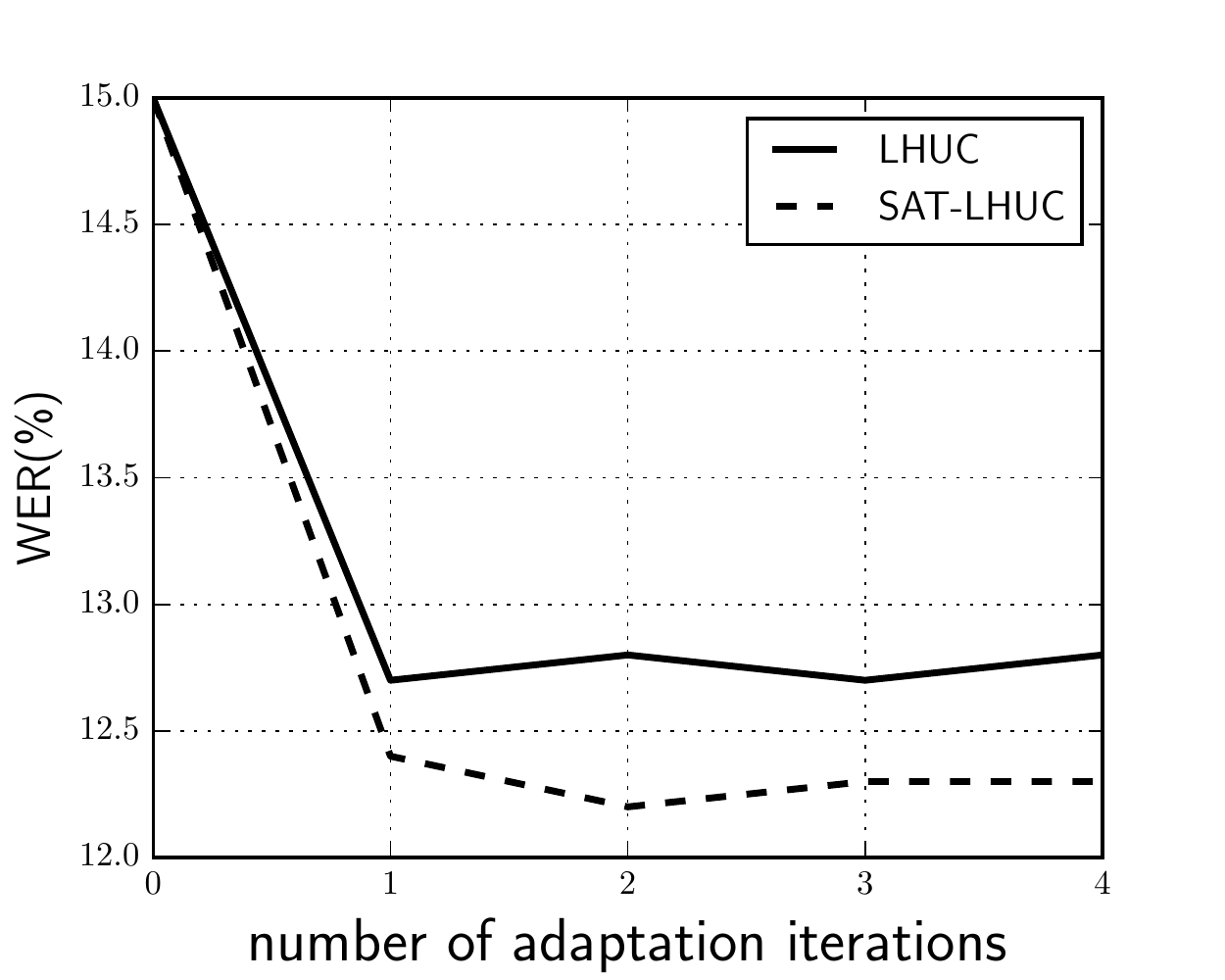}
} 
\subfigure[]{
  \includegraphics[width=0.33\textwidth]{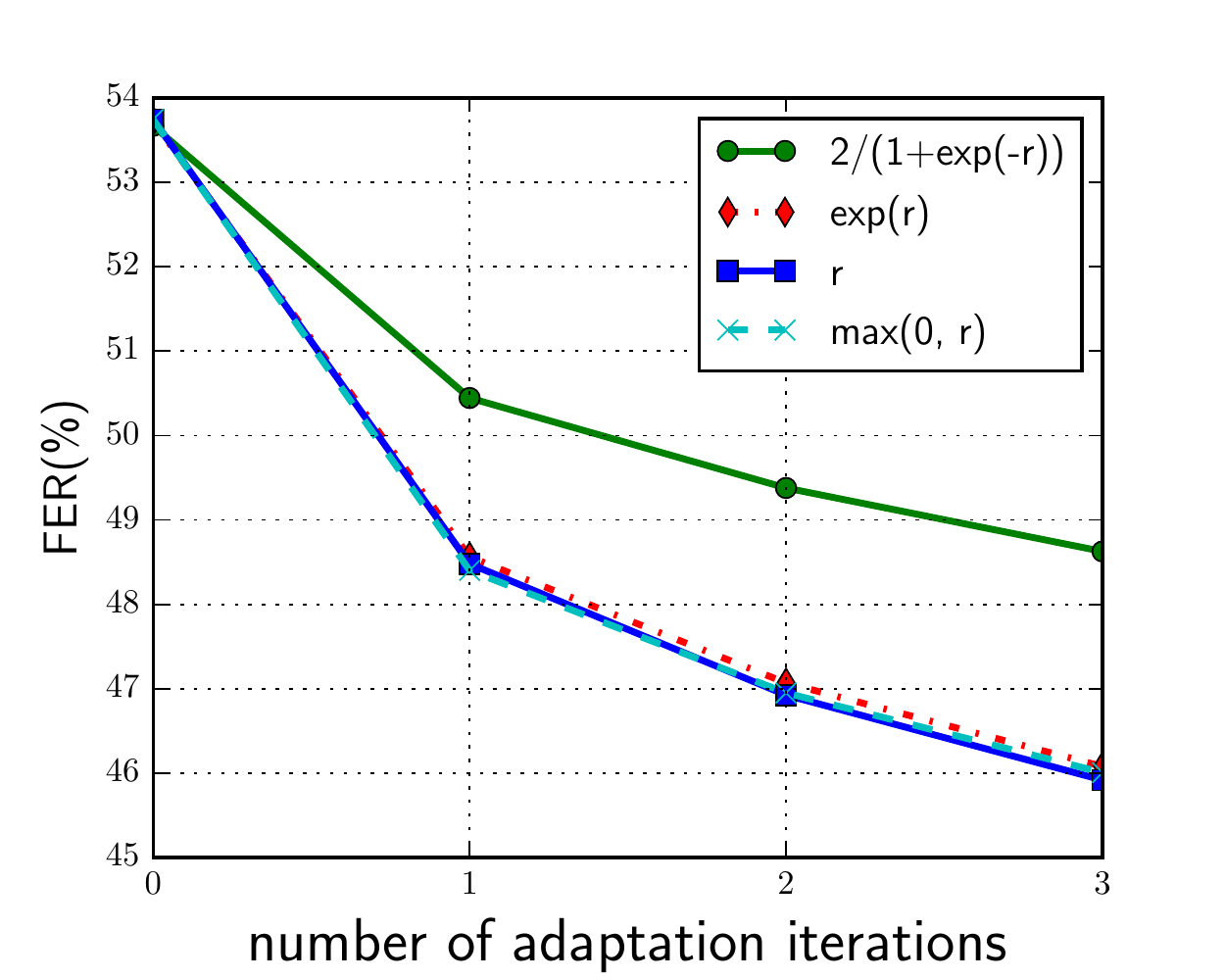}
} 
\vspace{-3mm}
\caption{WER(\%) on TED \tsta as a function of: a) number of adapted layers; and b) number of adaptation iterations; c) FER for re-parameterisation functions ($\xi$) used in adaptation.}
\label{fig:training_stats}
\end{figure*}

Our initial study concerned the hyper-parameters used with \lhuc adaptation. First, we used the TED talks to investigate how the word error rate (WER) is affected by adapting different layers in the model using \lhuc transforms.  The results, graphed in Fig.~\ref{fig:training_stats} (a), indicated that adapting only the bottom layer brings the largest drop in WER; however, adapting more layers further improves the accuracy for both \lhuc and \satlhuc approaches (adapting the other way round -- starting from the top layer -- is much less effective \cite{Swietojanski2014_lhuc}). Since obtaining the gradients for the $r$ parameters at each layer is inexpensive compared to the overall back-propagation, and we want to adapt at least the bottom layer, we apply \lhuc to each layer for the rest of this work.

Fig.~\ref{fig:training_stats} (b) shows WERs for the number of adaptation iterations. The results indicate that one sweep over the adaptation data (in this case \tsta) is sufficient and, more importantly, the model does not overfit when adapting with more iterations (despite the adaptation objective consistently improving -- Fig.~\ref{fig:training_stats} (c)). This suggests that it is not necessary to carefully regularise the model -- for example, by Kullback-Leibler divergence training \cite{Yu2013} which is usually required when adapting the weights of one or more layers in a network.

Finally, we explored how the form of the \lhuc re-parametrisation function $\xi$ affects the WER and frame error rate (FER) (Fig.~\ref{fig:training_stats} (c) and Table~\ref{tab:lhucreparm}). For test-only adaptation only a small WER  difference (0.1\% absolute) is observed, regardless of the large difference in frame accuracies. This supports our previous observation that \lhuc is robust against over-fitting. For \satlhuc training, a less constrained parametrisation was found to give better WERs for the SI model. Based on our control experiments, during \satlhuc training, setting $\xi$  to be the identity function (linear $r$) gave similar results to  $\xi(r)=\max(0,r)$ and $\xi(r)=\exp(r)$ and all were better than re-parametrising with $\xi(r)=2/(1+\exp(-r))$. This is expected as for full training the last approach constrains the range of back-propagated gradients. From now on, if not stated otherwise, we will use $\xi(r)=\exp(r)$ in the remainder of this paper.

We adapt our all models with the learning rate set to $0.8$ (regardless of $\xi(\cdot)$) and the basic training of both the SI and the \satlhuc models was performed with the initial learning rate set to $0.08$ and was later adjusted according to the \texttt{newbob} learning scheme \cite{Renals1992}.

\begin{table}[t]
\small
\caption{\label{tab:lhucreparm} WER(\%) for different re-parametrisation functions for \lhuc transforms on TED \tsta. Unadapted baseline WER is 15.0\%.}
\centerline{
\begin{tabular}{c|c|c|c}
$r$ & $2/(1+\exp(-r))$ & $\exp(r)$ & $\max(0, r)$ \\ 
\hline \hline
 12.8 & 12.8 & 12.7 & 12.7  \\ 
 \hline \hline
\end{tabular}}
\end{table}




\subsection {SAT-LHUC} \label{ssec:sat}

As described in section~\ref{sec:satlhuc}, \satlhuc training aims to regularise the hidden unit feature receptors so that they capture not just the average characteristics of training data, but also specific features of the different distributions the data was drawn from (for example, different training speakers). As a result, the model can be better tailored to unseen speakers by putting more importance to those units that were useful for training speakers with similar characteristics.

Prior to \satlhuc training we need to decide on how and which data should be used to estimate speaker-dependent and speaker-independent transforms. In this work we train \satlhuc models with frame-level~\cite{Swietojanski:ICASSP16}, segment-level and speaker-level clusters. For speaker- and segment-level transforms we decide which speakers or segments are going to be treated as SI or SD prior to training. For the frame-level \satlhuc approach, the SI/SD decisions are made separately for each data-point during training. In either scenario we ensure that the overall SD/SI ratio determined by $\gamma$ parameter is satisfied. The WER results for each of these three approaches ($\gamma=0.5$) are reported in Table~\ref{tab:satlevels}. Speaker-level \satlhuc training provides the highest WERs for both SI and SD decodes. Segment-level and frame-level \satlhuc training result in similar WERs for SI decodes, with a small advantage (0.1\% abs.) for the frame-level approach after adaptation.

\begin{figure}[ht!]
\center
  \includegraphics[width=0.95\columnwidth]{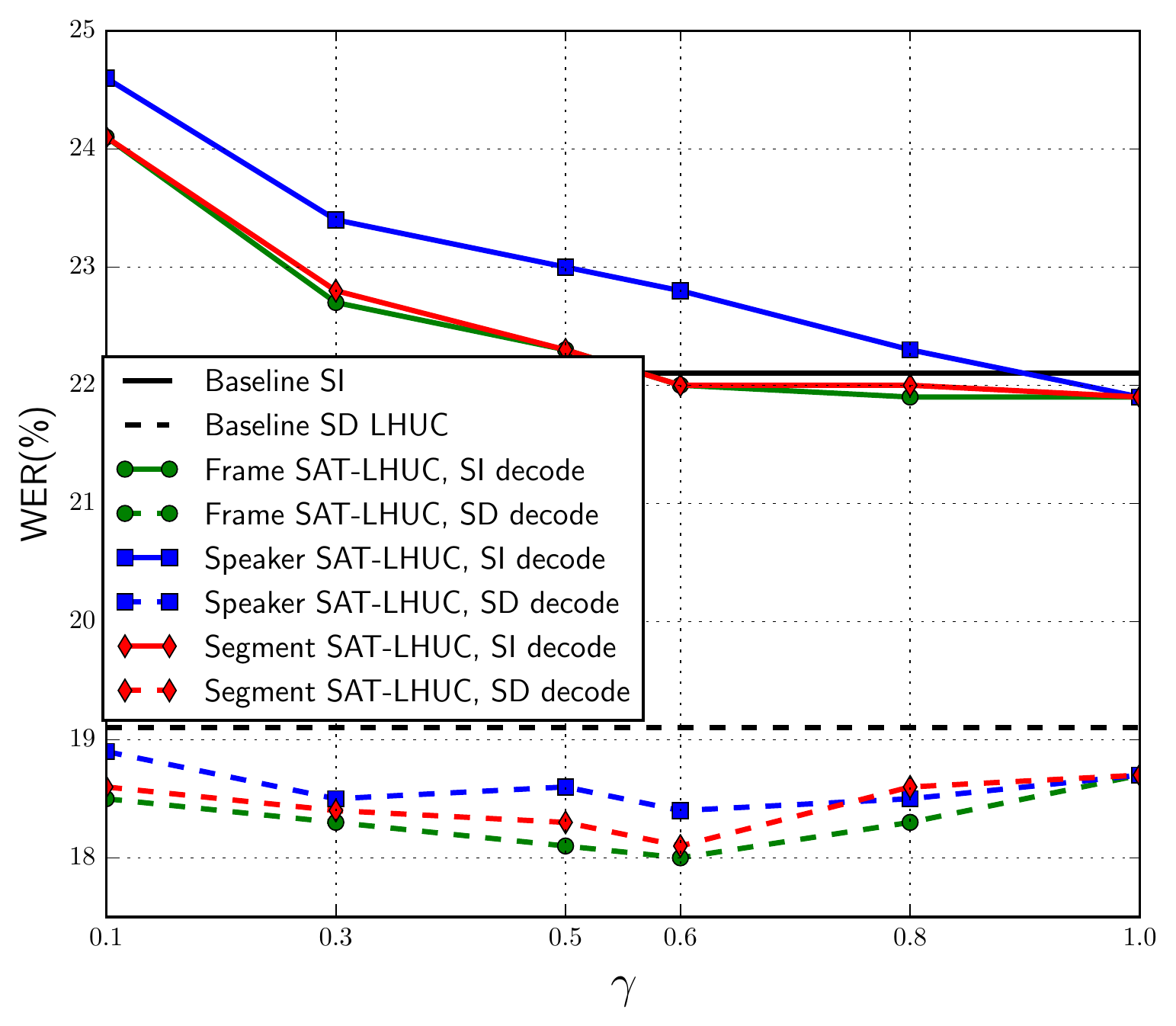}
\caption{WER(\%) for different sampling strategies \{per frame, per segment, per speaker\} 
for SAT-LHUC training and SI and SD decodes on TED \tstc.}
\label{fig:ted_as_p2}
\end{figure}

Fig.~\ref{fig:ted_as_p2} gives more insight on how the ratio of SI and SD data (determined by $\gamma$) affects the WER of the first-pass and adapted systems on TED \tstc.
The SI/SD split mainly affects the first pass accuracies with a substantial increase in SI WER when less than 30\% of the data is used to estimate the SI \lhuc transforms. However, once adapted, all variants obtained lower WERs compared to the baseline SI and \lhuc adapted model. For instance, when $\gamma=0.5$ the \satlhuc systems operating in SI mode obtained similar accuracies to the baseline SI model (22\%WER); however, the adapted \satlhuc model gave around 1\% absolute (6\% relative) decrease in WER compared with the SI baseline test-only adapted \lhuc model. The adaptation results for speaker-level \satlhuc training were worse by around 0.4\% absolute compared to segment- or frame-level \satlhuc training.  However, the difference, as shown experimentally in~\cite{Swietojanski:ICASSP16}, is mostly due to  poorer quality  adaptation targets resulting from the corresponding first pass \satlhuc systems rather than the differences in learned representations.  Managing a good trade-off between SI and SD ratios for \satlhuc is nevertheless an important aspect to take into account, and in our experience using around 50--60\% of data for the SI transform is a good task-independent setting. If different models for SI and SD decodes are acceptable, then further small gains in accuracy are observed~\cite{Swietojanski:ICASSP16}.

\begin{table}[t]
\small
\caption{\label{sat_strategies} WER(\%) for different sampling strategies and \satlhuc training (TED \tstc)}
\label{tab:satlevels}
\centerline{
\begin{tabular}{c|c||c|c|c}
& & \multicolumn{3}{c}{WER (\%) for sampling strategies} \\ \cline{3-5}
Model & Baseline & Per Speaker & Per Segment & Per Frame \\ 
\hline \hline
SI & 22.1 & 23.0 & 22.0 & 22.0  \\ 
SD & 19.1 & 18.6 & 18.1 & 18.0  \\ 
 \hline \hline
\end{tabular}}
\end{table}

We report the baseline \lhuc and \satlhuc comparisons on TED and AMI data in Tables ~\ref{tab:tedwers1} and ~\ref{tab:amiwers1}, respectively (further results, including a comparison to fMLLR transforms and on Switchboard data are in the next sections).  On TED (Table~\ref{tab:tedwers1}), \satlhuc models operating in SI mode ($\gamma=0.6$) have comparable WERs to SI models; however, adaptation resulted in a WER reduction of  0.3--1.1\% absolute (2--6\% relative) compared to test-only adaptation of the SI models. Similar results were observed on the AMI data (Table~\ref{tab:amiwers1}) where for both DNN and CNN models trained on FBANK features \lhuc adaptation decreased the WER by 2\% absolute (7\% relative) and \satlhuc training improved this result by 4\% relative for DNN models. As expected, the \satlhuc gain for CNNs was smaller when compared to DNN models, since the CNN layer can learn different patterns for different speakers which may be selected through the max-pooling operator at run-time.

\begin{table}[t]
\small
\caption{WER(\%) on TED talks (\tsta and \tstc).}
\label{tab:tedwers1}
\centerline{
\begin{tabular}{c|c||c|c}
  \multicolumn{2}{ c|| }{System}    & \multicolumn{2}{c}{IWSLT Test set} \\ \hline
  Training & Decoding   &  \tsta & \tstc  \\ 
\hline \hline
\multicolumn{4}{ l }{Baseline speaker-independent systems} \\ \hline
SI  & SI	     & 15.0 & 22.1  \\  
SAT-LHUC & SI    & 15.1 & 22.0  \\ 
\hline
\multicolumn{4}{ l }{Adapted systems} \\ \hline%
SI  & \LHUC 	 & 12.7 & 19.1 \\
\satlhuc & \LHUC  & 12.4 & 18.0 \\ 
\hline
\hline
\end{tabular}}
\vspace{-0.1cm}
\end{table}

\begin{table}[t]
\small
\caption{WER(\%) on AMI--IHM}
\label{tab:amiwers1}
\centerline{
\begin{tabular}{l|c||c|c}
Model & Features & \amidev & \amieval \\ 
\hline
DNN   	&  	FBANK & 26.8 		& 29.1   \\ 
+\lhuc 	&  	FBANK & 25.6 	& 27.1   \\ 
+\satlhuc & FBANK & 24.9        & 26.1   \\ \hline
CNN  &  	FBANK & 25.2 		& 27.1   \\ 
+\lhuc &  	FBANK & 24.3 		& 25.3   \\ 
+\satlhuc & FBANK & 23.9        & 24.8   \\
\hline \hline
\end{tabular}}
\vspace{-0.3cm}
\end{table}

\subsection {Sequence model adaptation} \label{ssec:sequence}

Model-based adaptation of sequence-trained DNNs (SE-DNN) is more challenging compared to adapting networks trained using cross-entropy: a mismatched adaptation objective (here cross-entropy) can easily erase sequence information from the weight matrices due to the well-known effect of catastrophic forgetting~\cite{French1999} in neural networks. Indeed Huang and Gong~\cite{Huang2015} report no gain from adapting SE-DNN models with cross-entropy adaptation objective and supervised adaptation targets. In those experiments, all weights in the model were updated and one needs to perform KL divergence regularised adaptation~\cite{Yu2013} or KL regularised sequence level adaptation to further improve on top of SE-DNN. It remains to be answered if one can get similar improvements using SE-DNN adaptation and first-pass transcripts.

In this work we adapt state-level minimum Bayes risk (sMBR) \cite{Kaiser2000, Kingsbury2009} sequence-trained models using \lhuc  and report  results on TED \tstb and \tstc in Table~\ref{tab:seqwers}. We kept all the \lhuc adaptation hyper-parameters the same as for CE models and obtained around 2\% absolute (11\% relative) WER reductions on \tstc for both SI and fMLLR SAT adapted SE-DNN systems. Interestingly, the obtained adaptation gain was similar to the cross-entropy models and \lhuc adaptation did not seem to disrupt the learned model's sequence representation.
\begin{table}[t]
\small
\caption{Summary of WER results of \lhuc adapted sequence models on TED \tstb and \tstc}
\label{tab:seqwers} 
\centerline{
\begin{tabular}{l||c|c}
Model & \tstb & \tstc \\ 
\hline \hline
DNN-CE    & 12.1 & 22.1  \\ \hline
DNN-sMBR    & 10.3 & 20.2  \\ 
+\lhuc      & 9.5 & 18.0 \\ \hline
+fMLLR      & 9.6 & 18.9 \\ 
++\lhuc 	& 8.9 & 15.8 \\ 
\hline \hline
\end{tabular}}
\end{table}

We compared our adaptation results to the most accurate system of the IWSLT--2013 TED transcription evaluation, which performed both feature- and model-space speaker adaptation~\cite{Huang2013_iwslt}. For model-space adaptation that system used a method which adapts DNNs with a speaker-dependent layer~\cite{Ochiai2014}. The results are reported in Table~\ref{tab:iwsltwers} where in the first block one can see a standard sequence-trained feature-space adapted system build from TED and 150 hours of out-of-domain data scoring 15.7\% WER, similar to the WER of our TED system (15.4\%), which also for IWSLT utilised 100 hours of out-of-domain AMI data. The 0.3\% difference could be explained by characteristics of the out-of-domain data used (\tstc is characterised by a large proportion of non-native speakers which is also typical for AMI data, hence benefits more our baseline systems).  When comparing both adaptation approaches operating in an unsupervised manner one can see that \lhuc gives much bigger improvements in WER compared to speaker-dependent layer, 2.1\% vs. 0.6\% absolute (14\% vs. 4\% relative) on \tstc. This allows our single-model system to match a considerably more sophisticated post-processing pipeline~\cite{Huang2013_iwslt}, as outlined in Table~\ref{tab:iwsltwers}. For less mismatched data (\tstb) adaptation is less important and our system has a WER 0.8\% absolute higher compared with the more sophisticated system.

From these experiments we conclude that \lhuc is an effective way to adapt sequence models in an unsupervised manner using a cross-entropy objective function, without the risk of removing learned sequence information.

\begin{table}[t]
\small
\caption{WERs for adapted sequence-trained models used in IWSLT evaluation.  Note, the results are not directly comparable to those reported on TED in Table~\ref{tab:seqwers} due different training data and feature pre-processing pipelines (see referenced papers for system details).}
\label{tab:iwsltwers}
\centerline{
\begin{tabular}{l||c|c}
Model & \tstb & \tstc \\ 
\hline \hline
\multicolumn{3}{l}{IWSLT2013 winner system (numbers taken from \cite{Huang2013_iwslt})} \\ \hline
DNN (sMBR) + HUB4 + WSJ & - & 15.7 \\ 
+ Six ROVER subsystems	& - & 14.8 \\ 
++ Automatic segmentation& - & 14.3 \\
+++ LM adapt. + RNN resc.        & - & 14.1 \\
+++++ SAT on DNN \cite{Ochiai2014}& 7.7 & 13.5 \\ \hline
\multicolumn{3}{l}{Our system \cite{Bell2014}} \\ \hline
DNN (sMBR) + AMI data  & 9.0 & 15.4 \\ 
+\lhuc 	& 8.5 & 13.3 \\ 
\hline \hline
\end{tabular}}
\end{table}


\subsection {Other aspects of adaptation} \label{ssec:quality}

\begin{figure}[ht!]
\center
  \includegraphics[width=1.\columnwidth]{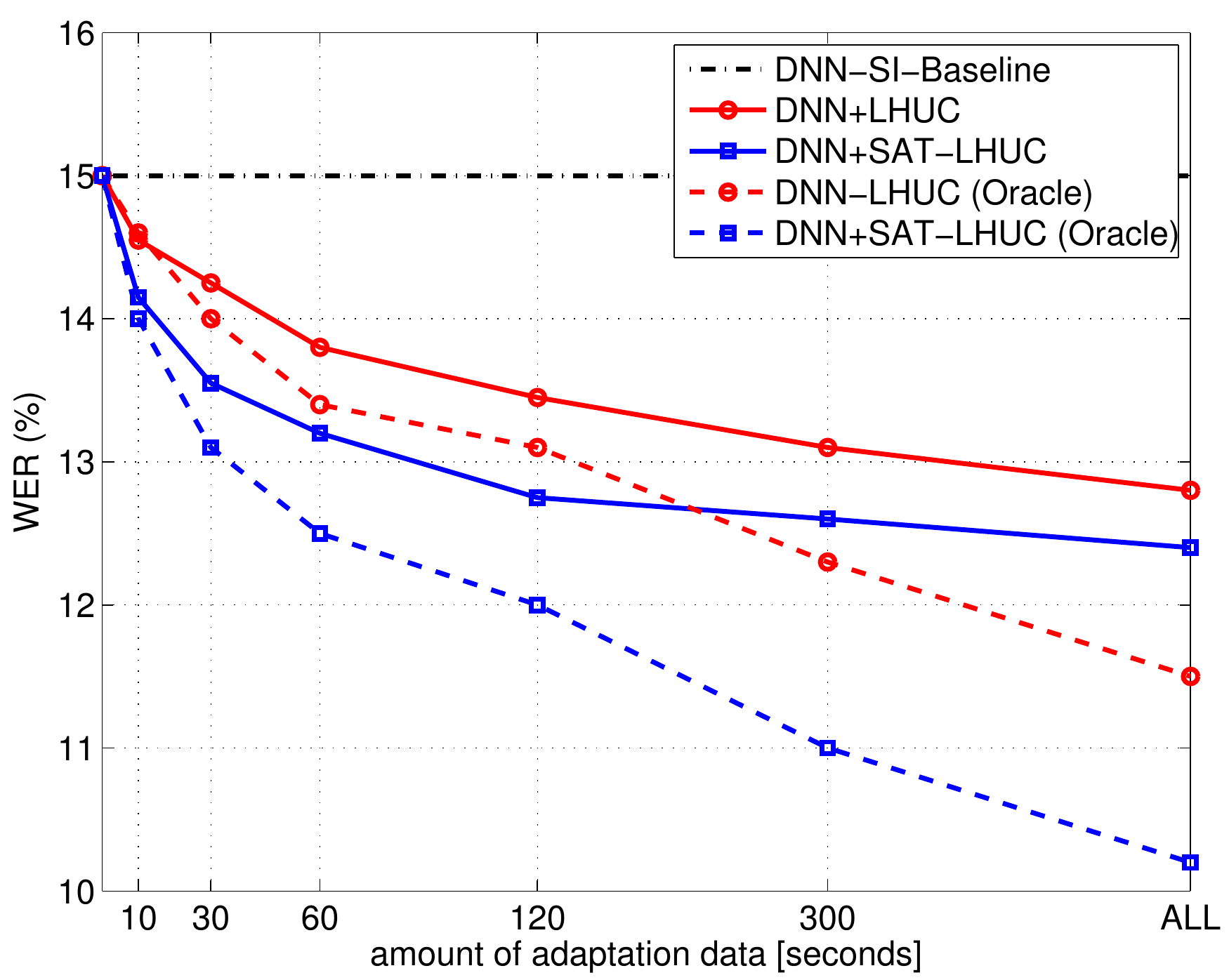}
\caption{WER(\%) for unsupervised and oracle adaptation data on TED \tsta.}
\label{fig:complementarity_lhuc}
\end{figure}

\textbf{Amount of adaptation data}: Fig~\ref{fig:complementarity_lhuc} shows the effect of the amount of adaptation data on WER for \lhuc and \satlhuc adapted models. As little as 10s of unsupervised adaptation data is already able to substantially decrease WERs (by 0.5--0.8\% absolute). The improvement for \satlhuc adaptation  compared with \lhuc is considerably larger -- roughly by a factor of two up to 30s adaptation data. As the duration of adaptation data increases the difference gets smaller; however \satlhuc results in consistently lower WERs than \lhuc in all cases (including full two pass adaptation).

We also investigated supervised (oracle) adaptation by aligning the acoustics with the reference transcriptions (dashed lines). Given supervised adaptation targets,  \lhuc and \satlhuc further substantially decrease WERs, with \satlhuc giving a consistent advantage over \lhuc. 

\begin{figure}[t]
  \includegraphics[width=1.0\columnwidth]{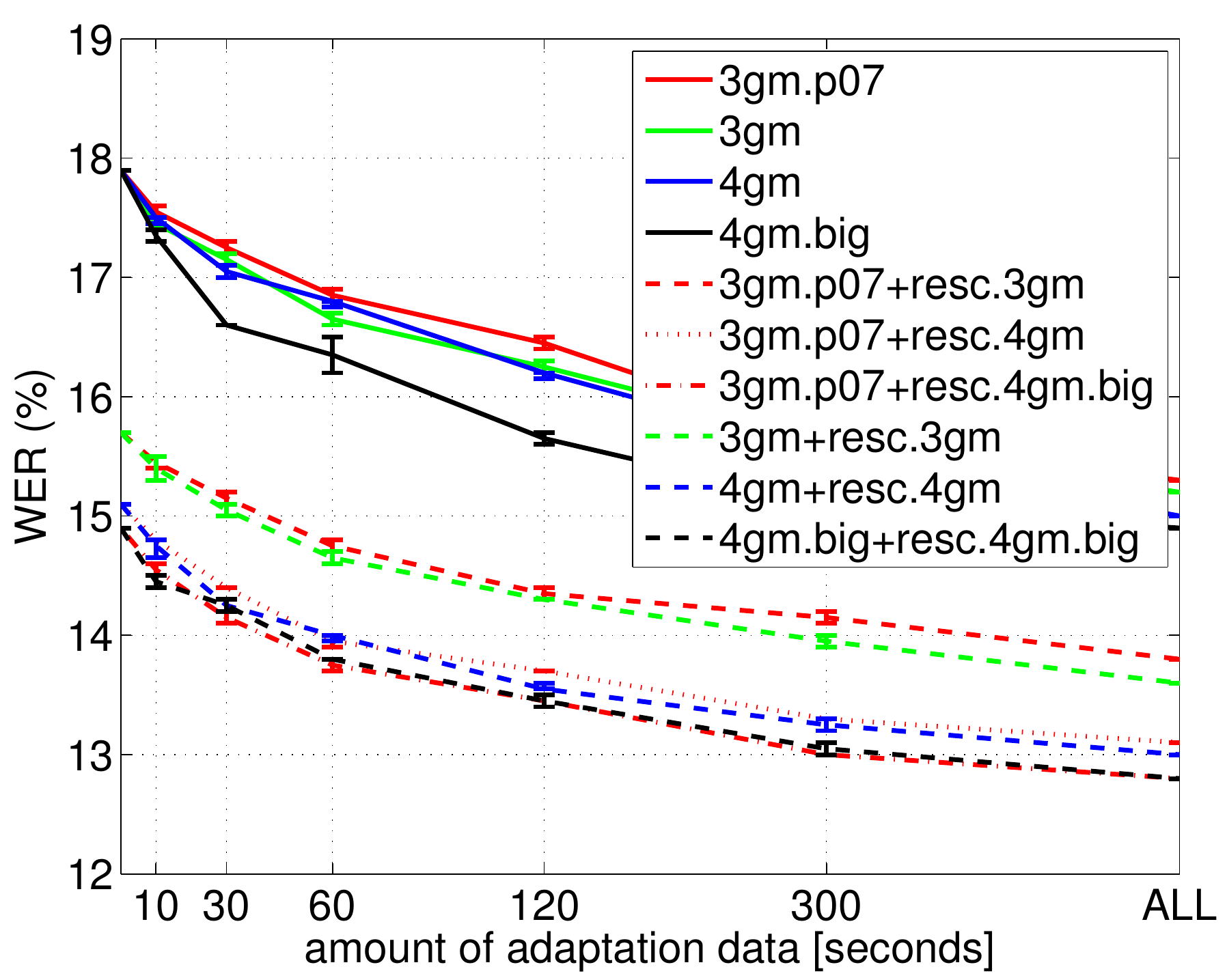}
\caption{WER(\%) for different qualities of adaptation targets on TED \tsta.}
\label{fig:as_data_quality}
\end{figure}

\textbf{Quality of adaptation targets}: Since our approach relies on a first-pass decoding, we investigated the extent to which \lhuc is sensitive to the quality of the adaptation targets. In this experiment we explored the differences resulting from different language models, and assumed that the first pass adaptation data was generated by either an SI or a \satlhuc model operating in SI mode. The main results are shown in Fig~\ref{fig:as_data_quality} where the solid lines show WERs obtained with a pruned 3-gram LM and different types of adaptation targets resulting from re-scoring the adaptation data with stronger LMs. One can see there is not much difference unless the adaptation data was re-scored with the largest 4-gram LM. This improvement diminishes in the final adapted system after re-scoring.  This suggests that the technique is not very sensitive to the quality of adaptation targets.  This trend holds regardless of the amount of data used for adaptation (ranging from 10s to several minutes per speaker). In related work~\cite{Miao2015} \lhuc was employed using alignments obtained from an SI-GMM system with a 8.1\% absolute higher WER than the corresponding SI DNN, and substantial gains were obtained over the unadapted SI DNN baseline -- although the WER reduction was considerably smaller (1\% absolute) compared to adaptation with alignments obtained with the corresponding SI DNN.

\textbf{Quality of data}: We also investigated how the quality of the acoustic data itself affects the adaptation accuracies, keeping the other ASR components fixed.  We performed an experiment on the AMI corpus using speech captured by individual headset microphones (IHM) and a single distant tabletop microphone (SDM). In case of IHM we adapt to the headset; in this experiment we assume we have speaker labels for the SDM data\footnote{In a real scenario for SDM data one would have to perform speaker diarisation in order to obtain speaker labels.}. The results are reported in  Table~\ref{tab:lhuc_ami_ihm_sdm}: \lhuc adaptation improves the accuracy in both experiments, although the gain for the SDM condition is smaller; however, the SDM system is characterised by twice as large WERs. Notice that \lhuc has also been successfully applied to channel normalisation between distant and close talking microphones~\cite{himawan2015towards}.

\begin{table}[t]
\small
\caption{WER(\%) on AMI--IHM and AMI--SDM using adapted CNNs.}
\label{tab:lhuc_ami_ihm_sdm}
\centerline{
\begin{tabular}{l||c|c}
Model & \amidev & \amieval \\ 
\hline \hline
CNN (IHM) & 25.2 		& 27.1   \\ 
+\lhuc &  24.3 		& 25.3   \\ \hline
CNN (SDM) & 49.8 		& 54.4   \\ 
+\lhuc &  48.8		& 53.1   \\ 
\hline \hline
\end{tabular}}
\end{table}

\textbf{One-shot adaptation}: By one-shot adaptation we mean the scenario in which \lhuc transforms were estimated once for a held-out speaker and then used many times in a single pass system for this speaker.  We performed those experiments on AMI IHM data, and report results on \amidev and \amieval which contain 21 and 16 unique speakers taking part in 18 and 16 different meetings, respectively. Each speaker participates in multiple meetings: to some degree, adapting to a speaker in one meeting, then applying the adaptation transform to the same speaker in the other meetings simulates a real-life condition where it is possible to assume the speaker identity without having to perform speaker diarisation (e.g. personal devices).
The results of this experiment (Table~\ref{tab:oneshot}) indicate that \lhuc retains the accuracies of two-pass systems by providing almost identical results when comparing \lhuc estimated in a full two-pass system and when the unsupervised transforms are re-used in the \lhuc.\texttt{one-shot} experiment.

\begin{table}[t]
\small
\caption{WER(\%) on AMI--IHM and one-shot adaptation}
\label{tab:oneshot}
\centerline{
\begin{tabular}{l||c|c}
Model & \amidev & \amieval \\ 
\hline \hline
CNN &  	25.2 		& 27.1   \\ 
+\lhuc &  	24.3 		& 25.3   \\ 
+\lhuc.\texttt{one-shot}  & 24.3 		& 25.4   \\
\hline \hline
\end{tabular}}
\end{table}

\subsection{Complementarity to feature normalisation} \label{ssec:complementarity}
Feature-space adaptation using fMLLR is a very reliable current form of speaker adaptation, so it is of great interest to explore how complementary the proposed approaches are to SAT training with fMLLR transforms.\footnote{Due to space constraints we do not make an explicit comparisons to other techniques such as auxiliary i-vector features or speaker-codes; however, the literature suggest that the use of i-vectors give similar \cite{Saon2013} results when compared to fMLLR trained models. Related recent studies also show \lhuc is at least as good as the standard use of i-vector features \cite{Miao2015, Samarakoon:ICASSP16}.}

\begin{figure}[ht!]
\center
  \includegraphics[width=1.0\columnwidth]{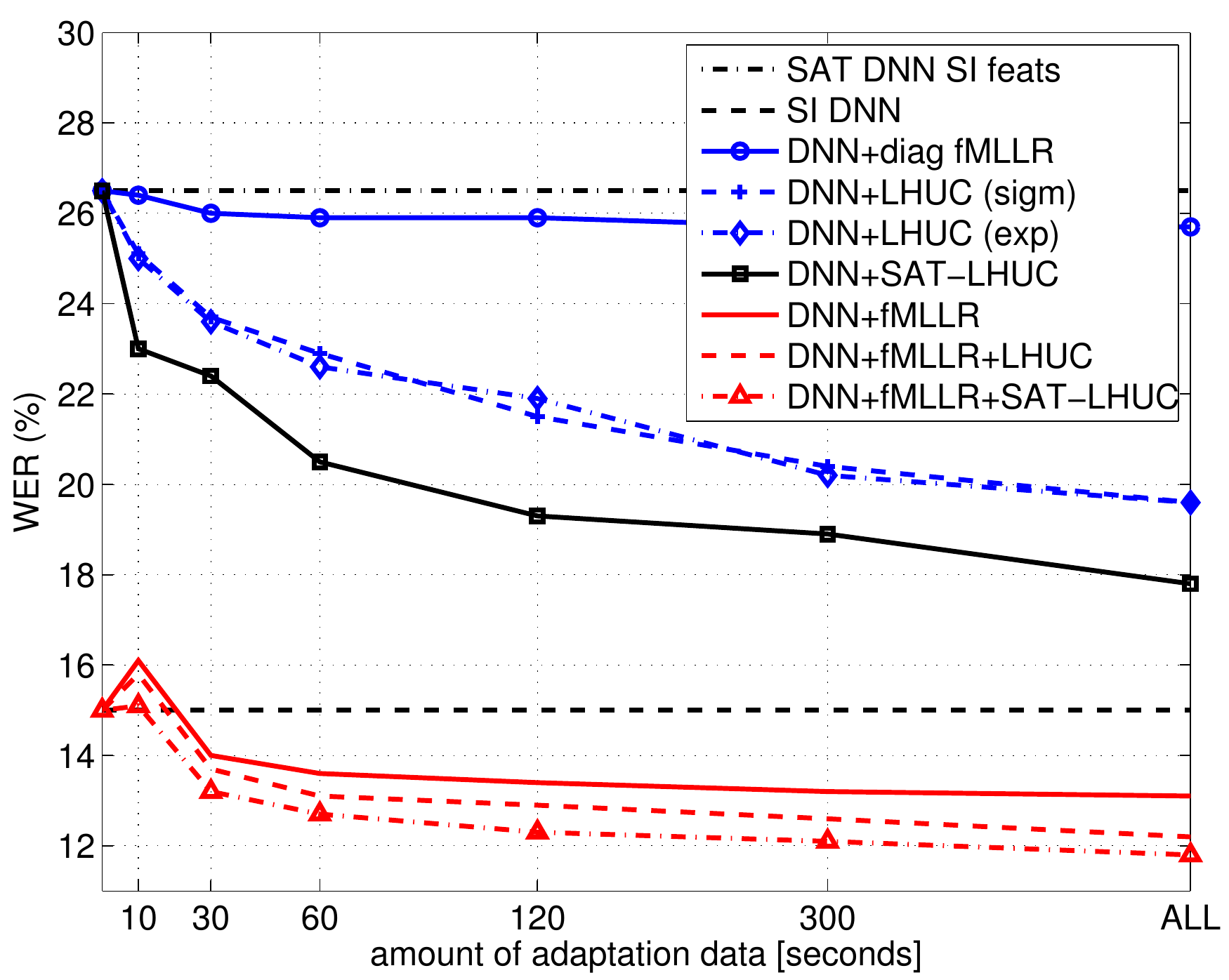}
\caption{WER(\%) for \lhuc, \satlhuc, and SAT-fMLLR (and combinations) on TED \tsta.}
\label{fig:complementarity_fmllr}
\end{figure}

We compared \lhuc and \satlhuc to SAT-fMLLR training using TED \tsta (Fig~\ref{fig:complementarity_fmllr}, red curves).  We also compared both techniques, including a comparison in terms of the amount of data used to estimate each type of transform.  fMLLR transforms estimated on 10s of unsupervised data result in an increase in WER compared with the SI-trained baseline (16.1\% vs. 15.0\%).  When combined with \lhuc or \satlhuc some of this deterioration  was recovered (similar results using  \lhuc alone were reported in Fig~\ref{fig:complementarity_lhuc}). For more adaptation data (30s or more) fMLLR improved the accuracies by around 1--2\% absolute and  combination with \lhuc (or \satlhuc) resulted in an additional 1\% reduction in WER (see also Table~\ref{tab:tedwers} in the next section for further results).

We also investigated (in a rather unrealistic experiment) how much mismatch in feature space one can normalise in model space with \lhuc. To do so, we used a SAT-fMLLR trained model with unadapted PLP features which gave a large increase in WER (26\% vs 15\%). Then, using unsupervised adaptation targets obtained from the feature-mismatched decoding both \lhuc and \satlhuc were applied. The results (also presented in Fig.~\ref{fig:complementarity_fmllr}) indicate that a very large portion of the WER increase can be effectively compensated in model space -- more than 8\% absolute.  As found before, test-only re-parametrisation functions ($\exp(r)$ vs. $2/(1+\exp(-r))$) have negligible impact on the adaptation results, and \satlhuc again provides better results.

\subsection{Adaptation Summary} \label{ssec:summary}

In this section we summarise our results, applying \lhuc and \satlhuc to TED, AMI, and Switchboard.  Table~\ref{tab:tedwers} contains results for four IWSLT test sets (\dev, \tsta, \tstb, and \tstc): in most scenarios \satlhuc results in a lower WER than \lhuc and both techniques are complementary with SAT-fMLLR training.  

Similar conclusions can be drawn from experiments on AMI (Table~\ref{tab:amiwers}) where \lhuc and \satlhuc were found to effectively adapt DNN and CNN models trained on FBANK features. \satlhuc trained DNN models gave the same final results as the more complicated SAT-fMLLR+\lhuc system.

\begin{table}[t]
\small
\caption{WER (\%)  on various TED development and test sets from IWSLT12 and IWSLT13 evaluations.}
\label{tab:tedwers}
\centerline{
\begin{tabular}{l||c|c|c|c}
Model & \dev & \tsta & \tstb & \tstc \\ 
\hline \hline
DNN			& 15.4 & 15.0  & 12.1 & 22.1  \\ 
+\lhuc      & 14.5 & 12.8  & 10.9 & 19.1 \\ 
+\satlhuc   & 14.0 & 12.4  & 10.9 & 18.0 \\ \hline
+fMLLR      & 14.5 & 12.9  & 10.9 & 20.8 \\ 
++\lhuc     & 14.1 & 11.8  & 10.3 & 18.4 \\ 
++\satlhuc   & 13.7   & 11.6  & 9.9 & 17.6 \\ 
\hline \hline
\end{tabular}}
\end{table}

\begin{table}[t]
\small
\caption{WER(\%) on AMI--IHM}
\label{tab:amiwers}
\centerline{
\begin{tabular}{l|c||c|c}
Model & Features & \amidev & \amieval \\ 
\hline \hline
DNN  	&  	FMLLR & 26.2 		& 27.3   \\ 
+\lhuc  	&  	FMLLR & 25.6 		& 26.2   \\
\hline 
DNN   	&  	FBANK & 26.8 		& 29.1   \\ 
+\lhuc 	&  	FBANK & 25.6 	& 27.1   \\ 
+\satlhuc & FBANK & 24.9        & 26.1   \\ \hline
\hline
\end{tabular}}
\end{table}

On Switchboard, in contrast to other corpora, we observed that test-only LHUC does not match the WERs obtained from SAT-fMLLR models (Table~\ref{tab:swbd_wers}).  The SI system has a WER of 21.7\% compared with 20.7\% for the test-only LHUC and 20.2\% for the SAT-fMLLR system.  The improvement obtained using test-only LHUC is comparable to that obtained with other test-only adaptation techniques, e.g. feature-space discriminative linear regression (fDLR) \cite{Seide2011}, but neither of these matches  SAT trained feature transform models.  This could be due to the fact Switchboard data is narrow-band and as such contains less information for discrimination between speakers~\cite{Wester_IS2015}, especially when estimating relevant statistics from small amounts of unsupervised adaptation data.  Another potential reason could be related to the fact that the Switchboard part of $\swbdeval$ is characterised by a large overlap between training and test speakers -- 36 out of 40 test speakers are observed in training \cite{fiscus2000}, which limits the need for adaptation, but also enables models to learn much more accurate speaker-characteristics during supervised speaker adaptive training.

Adaptation using \satlhuc (20.3\% WER) almost matches SAT-fMLLR (20.2\%). We also observe that \LHUC performs relatively better under more mismatched conditions (the Callhome (CHE) subset of \swbdeval), similar to what we observed on TED. 

\begin{table}[t]
\small
\caption{WER(\%) on Switchboard \swbdeval.}
\label{tab:swbd_wers}
\centerline{
\begin{tabular}{l||c|c|c}
      & \multicolumn{3}{c}{\swbdeval} \\ \cline{2-4}
Model & SWB & CHE & TOTAL \\ 
\hline \hline
DNN 	& 15.2 	 & 28.2 		&  21.7 \\
+\lhuc  	&  	14.7 & 26.6    & 20.7 \\ 
++\satlhuc  	&  	14.6 &  25.9   & 20.3 \\ \hline
+fMLLR  	  & 14.2 & 26.2 	& 20.2  \\
++\lhuc 	 &  14.2 & 25.6 & 19.9  \\
++\satlhuc   & 14.1 & 25.6  & 19.9 \\
 \hline \hline
\end{tabular}}
\end{table}

\begin{figure*}[ht!]
\center
  \includegraphics[width=1.0\textwidth]{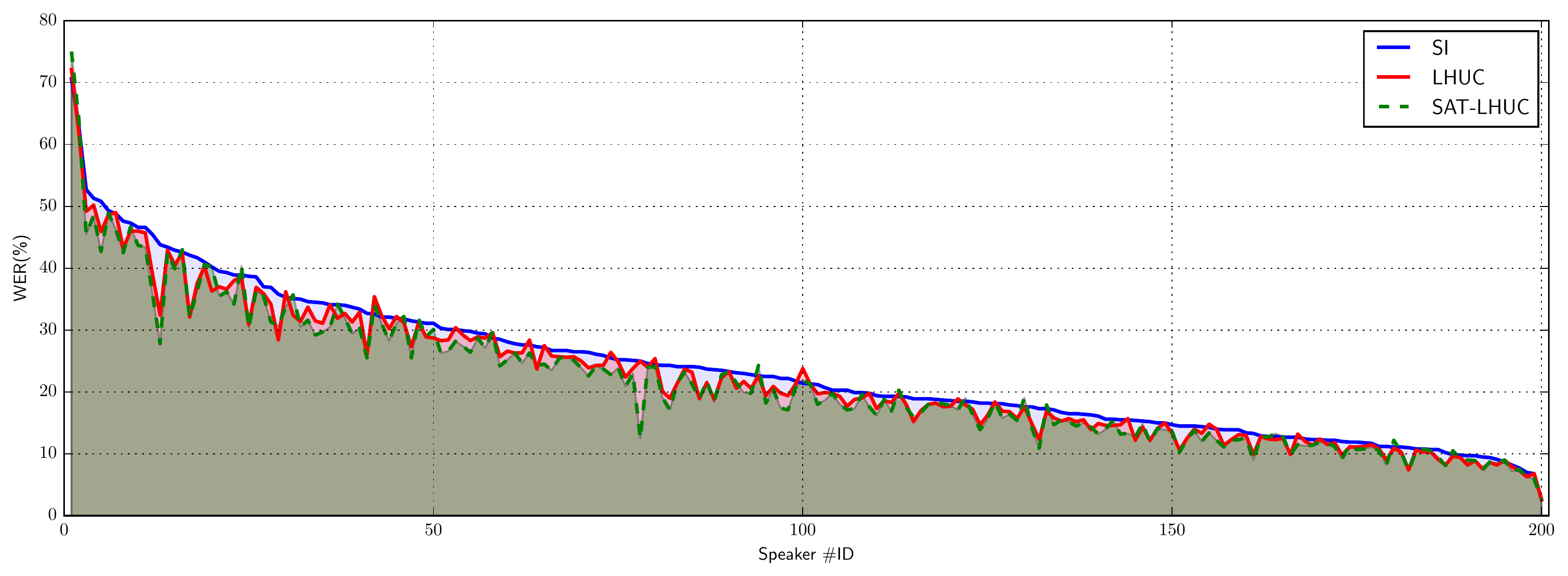}
\caption{Summary of WERs(\%) obtained with \lhuc and \satlhuc adaptation techniques on test speakers of TED, SWBD and AMI corpora (results are sorted in descending WER order for the SI system). For \lhuc the average observed improvement per speaker was at 1.6\% absolute (7.0\% relative). The same statistic for \satlhuc was at 2.3\% absolute (9.7\% relative). The maximum observed WER decrease per speaker was 11.4\% absolute (32.7\% relative) and 16.0\% absolute (50\% relative) for \lhuc and \satlhuc, respectively. WERs decreased for 89\% of speakers using \LHUC adaptation.}
\label{fig:summary}
\end{figure*}

Finally, in Fig~\ref{fig:summary} we show the WERs obtained for  200 speakers across the TED, AMI, and SWBD test sets.  We observe that for 89\% of speakers \lhuc and \satlhuc adaptation reduced the WER, and that \satlhuc gives a consistent reduction over \lhuc.

\section{LHUC for Factorisation} \label{sec:factorisation}

\def \R {\mathbf r}
\def \Rhat {\mathbf {\hat{r}}}
\def \b #1{\textbf{#1}}
\def \f {\phi}
\def \nlf #1{\f^{#1}}

We  applied \lhuc to adapt to both the speaker and the acoustic environment.  If multi-condition data is available for a speaker, then it is possible to define a set of joint speaker-environment \lhuc transforms.  Alternatively, we can estimate two set of transforms -- for speaker $\R_S$ and for environment $\R_E$ -- and then linearly interpolate them, using hyper-parameter $\alpha$, to derive a combined transform $\Rhat_{SE}$ as follows:
\begin{equation} \label{eq:intrp}
\xi \left ( \Rhat^l_{SE} \right ) = \alpha \xi \left ( \R^l_S \right ) + (1-\alpha) \xi \left ( \R^l_E \right )
\end{equation} 
Notice, that although both types of transforms are estimated in an unsupervised manner we assume that the test environment is known, allowing the correct environmental transform to be selected. This adaptation to the test environment is similar to that of Li et al~\cite{jinyu2014factor}.

We adapted baseline multi-condition trained DNN models~\cite{Seltzer2013}  to the speaker ($\R_S$) and the environment ($\R_E$).  The $\R_S$ transforms were estimated only on \emph{clean} speech; similarly the environment transforms were estimated for each scenario (one set of $\R_E$ per scenario) using multiple speakers (hence, we have 7 different environmental transforms).  To avoid learning joint speaker-environment transforms the target speaker's data was removed from environment adaptation material (e.g. when estimating transforms for the \texttt{restaurant} environment, we use all restaurant data except the one for the target speaker). 

The results (Table~\ref{tab:mdnn}) show that both standalone speaker or environment adaptation \lhuc adaptation improve over an unadapted system (13.1\%($S$) and 13.3\%($E$) vs. 13.9\%) but, as expected, a single transform estimated jointly on the target speaker and environment has a lower WER (12.4\%). However, when interpolated with $\alpha=0.7$ the result of the factorised model improves to 12.7\% WER, although still higher that joint estimation.  However, adaptation data for joint speaker-environment adaptation is not available in many scenarios, and the factorised adaptation based on interpolation is more flexible.

\begin{table} 
\caption{Results on Aurora 4. Multi-condition DNN model.}
\label{tab:mdnn}
\center{
\begin{tabular}{ l | c | c | c | c | c }
  Model & A & B & C & D & AVG \\ \hline \hline
  DNN 		& 5.1 & 9.3 & 9.3 & 20.8 & 13.9\\
  DNN + $\R_S$ & 4.3 & 9.3 & 6.9 & 19.3 & 13.1 \\
  DNN + $\R_E$ & 5.0     & 9.0   & 8.5   & 19.8  & 13.3 \\
  DNN + $\R_{SE\;JOINT}$  & 4.5  & 8.6   & 7.4     & 18.3  & 12.4 \\ \hline
  DNN + $\Rhat_{SE}$, $\alpha=0.5$ & 4.6  & 8.9   & 7.7    & 19.1  & 12.9 \\
  DNN +  $\Rhat_{SE}$, $\alpha=0.7$ & 4.5     & 8.8   & 7.2   & 18.9  & \b{12.7} \\ 
  \hline \hline
\end{tabular}
}
\end{table}

We also trained more competitive models following Rennie et al~\cite{Rennie2014}:  Maxout~\cite{Goodfellow2013} CNN models were trained using annealed dropout~\cite{srivastava2014dropout}. In this work we used alignments obtained by aligning a corresponding multi-condition model as ground-truth labels, rather than replicating clean alignments to multi-condition data, in contrast to \cite{Rennie2014}: this is likely to explain differences in the reported baselines (10.9\% compared with 10.5\% in \cite{Rennie2014}). The results for the joint optimisation are reported in  Table~\ref{tab:maxnn} where one can notice large improvements with unsupervised \lhuc adaptation.

\begin{table}
\caption{Results on Aurora 4. Multi-condition Maxout-CNN model, with and without annealed dropout (AD).}
\label{tab:maxnn}
\center{
\begin{tabular}{ l | c | c | c | c | c }
  Model & A & B & C & D & AVG \\ \hline \hline
  MaxCNN & 4.2     & 7.7   & 7.9   & 17.4  & 11.6 \\ 
  MaxCNN +  $\R_{SE\;JOINT}$ & 3.7 & 6.3   & 5.5   & 14.3  & 9.5 \\ \hline
  AD MaxCNN & 4.3     & 7.7   & 7.2   & 15.6  & 10.9 \\ 
  AD MaxCNN +  $\R_{SE\;JOINT}$ & 3.4 & 5.7   & 6.1   & 13.4  & 8.7 \\ \hline \hline
\end{tabular}
}
\vspace{-0.1cm}
\end{table}

\begin{figure}[!hbpt]
\centering
\subfigure[]{
  \includegraphics[width=0.82\columnwidth]{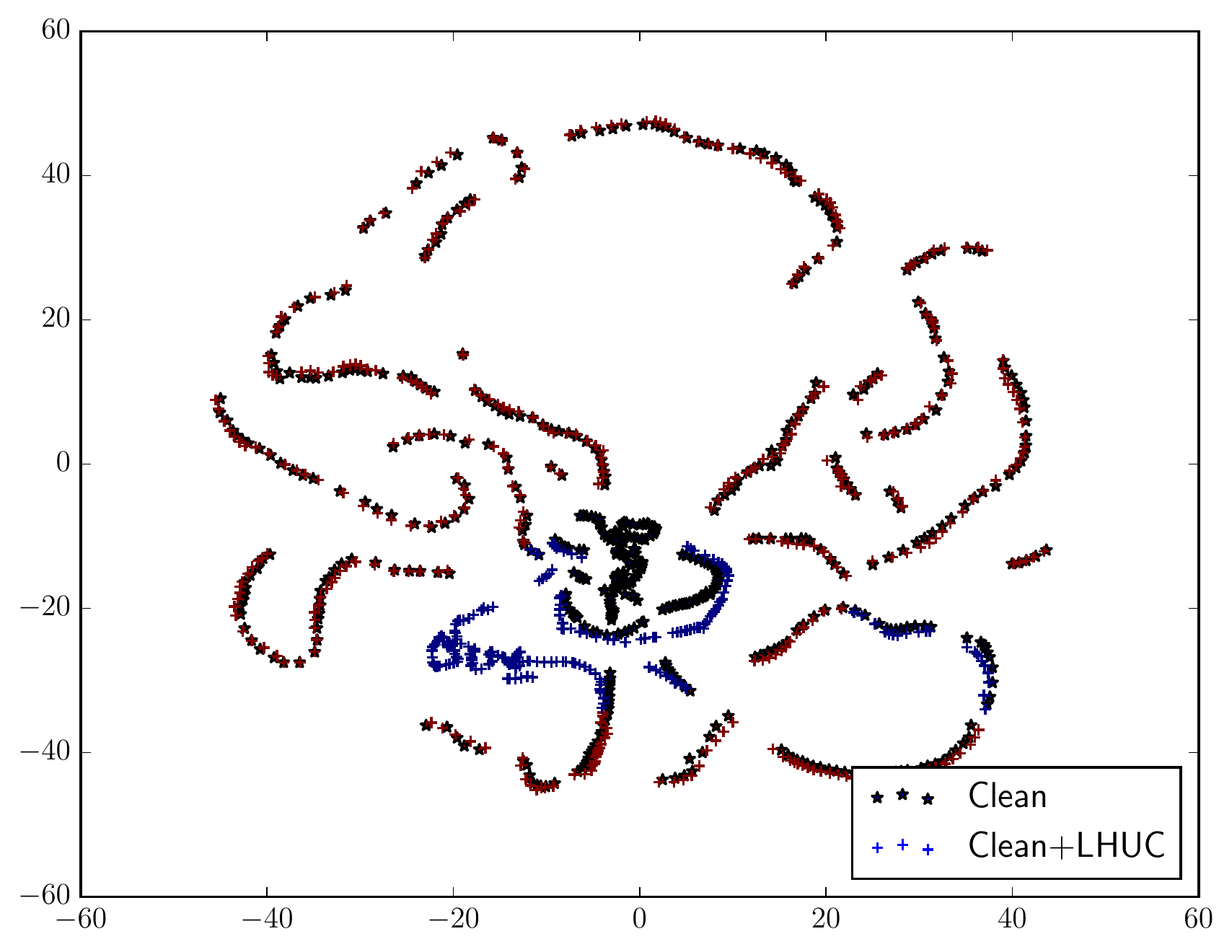}
}
\vspace{-3mm}
 \subfigure[]{
   \includegraphics[width=0.82\columnwidth]{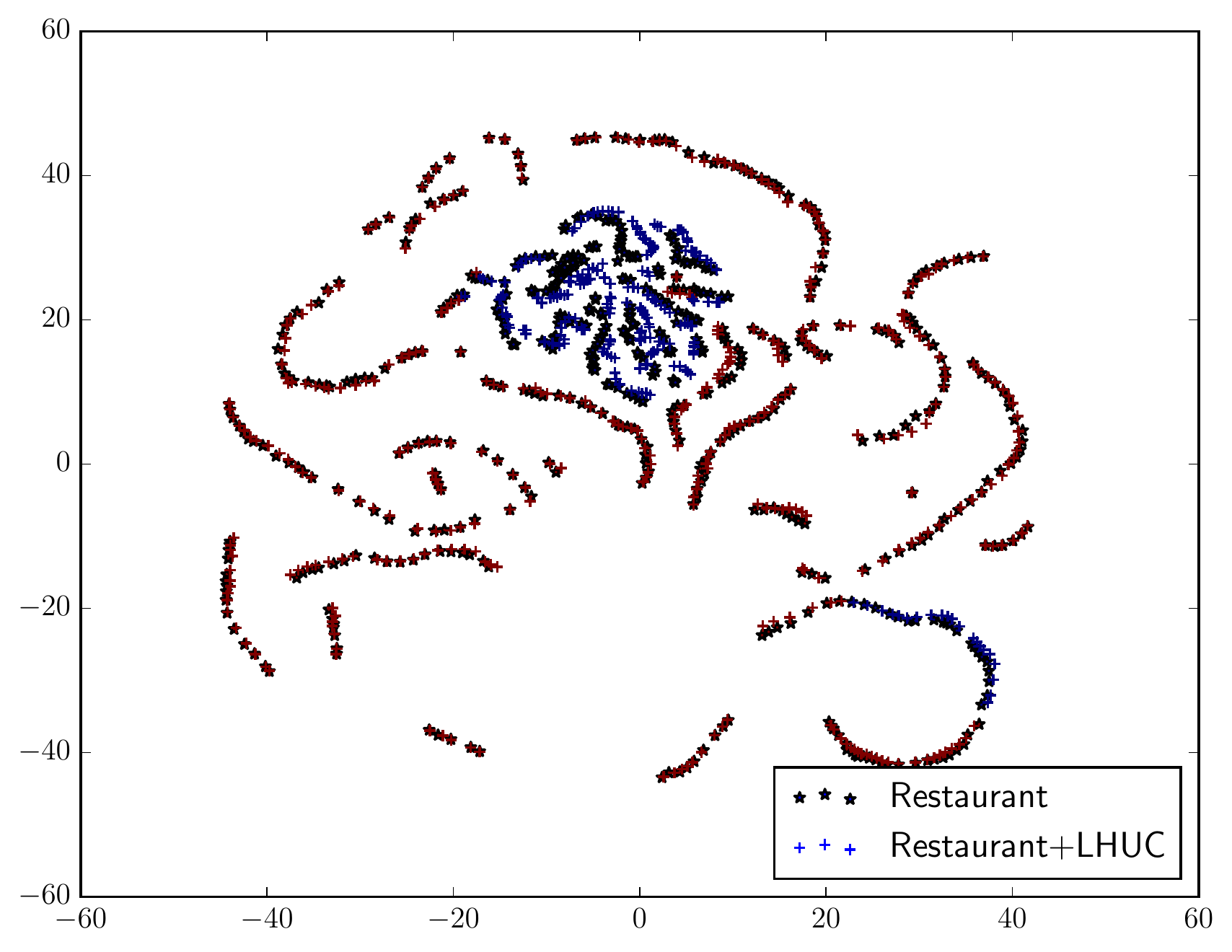}
 } 
\caption{tSNE plots (best viewed in color) of the top hidden layer before and after adaptation for an utterance recorded in (a) clean and (b) noisy (restaurant) environment, using the annealed dropout maxout CNN. The model can normalise the phonetic space between conditions (brown color), keeping two different spaces for non-speech frames (blue color) under clean and noisy conditions. The effect of \lhuc is mostly visible for non-speech frames.}
\label{fig:tsnea4}
\vspace{-0.3cm}
\end{figure}

Finally, we  visualise the top hidden layer activations of the annealed dropout Maxout CNN using stochastic neighbourhood embedding (tSNE)~\cite{Maaten2008_tsne} for one utterance recorded under clean and noisy (restaurant) conditions (Fig.~\ref{fig:tsnea4}). 

\section{Conclusions}

We have presented the \LHUC approach to unsupervised adaptation of neural network acoustic models in both test-only (\lhuc) and SAT  (\satlhuc) frameworks, evaluating them using four standard speech recognition corpora: TED talks as used in the IWSLT evaluations, AMI, Switchboard, and Aurora4.  Our experimental results indicate that both \lhuc and \satlhuc can provide significant improvements in word error rates (5--23\% relative depending on test set and task).  \LHUC adaptation works well unsupervised and with small amounts of data (as little as 10s),  is complementary to feature space normalisation transforms such as SAT-fMLLR, and can be used for unsupervised adaptation of sequence-trained DNN acoustic models using a cross-entropy adaptation objective function.  Furthermore we have demonstrated that it can be applied in a factorised way, estimating and interpolating separate transforms for adaptation to the acoustic environment and speaker.

\ifCLASSOPTIONcaptionsoff
  \newpage
\fi

\bibliographystyle{IEEEbib}
\bibliography{master}

\begin{IEEEbiography}[{\includegraphics[width=1.1in,height=2.25in,clip,keepaspectratio]{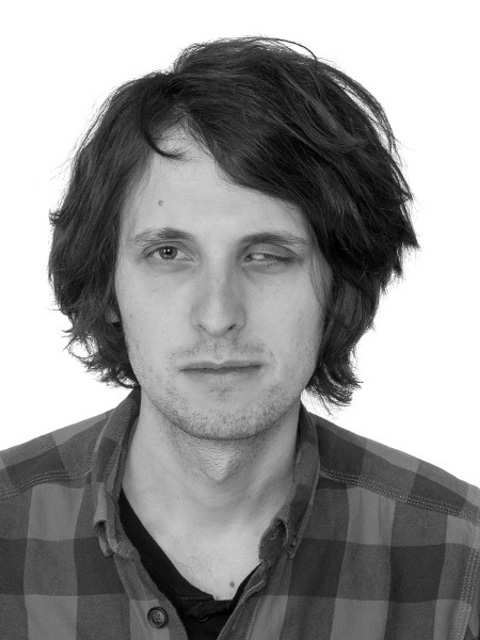}}]{Pawel Swietojanski} received his M.Sc. degree from AGH University of Science and Technology in Cracow, Poland and is now a Ph.D. candidate in Informatics at the Centre for Speech Technology Research, School of Informatics, University of Edinburgh, UK. His main research interests are in machine learning and its applications to speech processing, with a particular focus on learning representations for acoustic modelling in speech recognition.
\end{IEEEbiography}

\begin{IEEEbiography}[{\includegraphics[width=1.1in,height=2.25in,clip,keepaspectratio]{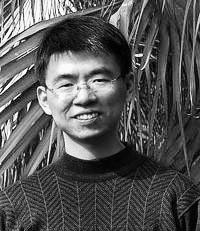}}]{Jinyu Li} (M'08) received the Ph.D. degree from Georgia Institute of Technology, U.S. From 2000 to 2003, he was a Researcher at Intel China Research Center and a Research Manager at iFlytek, China. Currently, he is a Principal Applied Scientist at Microsoft, working as a technical lead to design and improve speech modeling algorithms and technologies that ensure industry state-of-the-art speech recognition accuracy for Microsoft products. His major research interests cover several topics in speech recognition and machine learning, including noise robustness, deep learning, discriminative training, and feature extraction. He has authored one book and over 60 papers, and awarded over 10 patents.
\end{IEEEbiography}

\begin{IEEEbiography}[{\includegraphics[width=1.1in,height=2.25in,clip,keepaspectratio]{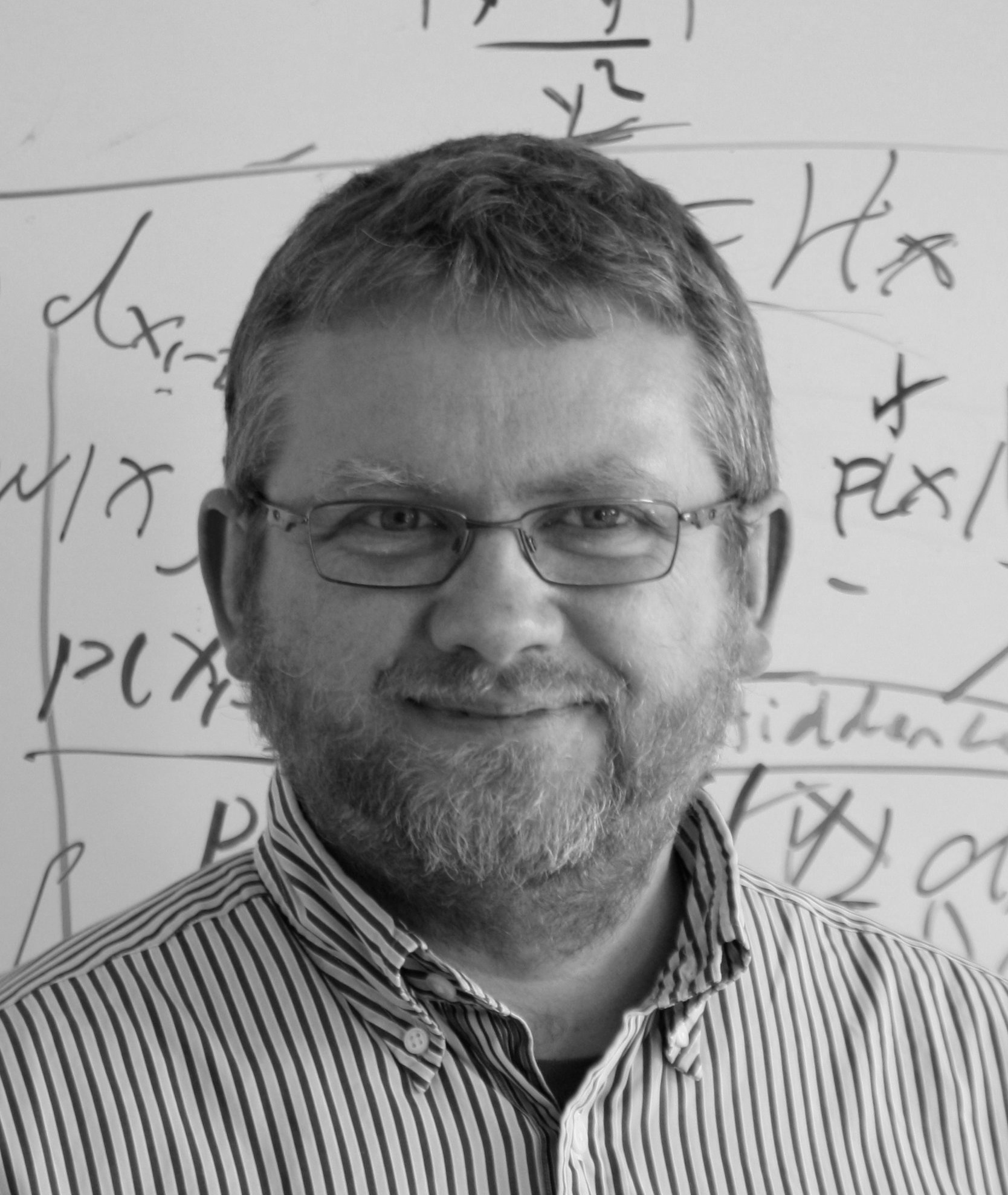}}]{Steve Renals} (M'91 --- SM'11 -- F'14) is professor of speech technology at the University of Edinburgh.  He received a BSc from the University of Sheffield and an MSc and PhD from Edinburgh.  He has previously had positions at ICSI Berkeley, the University of Cambridge, and the University of Sheffield.  His research interests are in speech and language processing.
\end{IEEEbiography}

\end{document}